\documentclass{article}
\usepackage{PRIMEarxiv}
\usepackage[shortlabels]{enumitem}
\usepackage[title]{appendix}
\usepackage[numbers]{natbib}
\usepackage[utf8]{inputenc} 
\usepackage[T1]{fontenc}    
\usepackage{hyperref}       
\usepackage{url}            
\usepackage{booktabs}       
\usepackage{amsfonts}       
\usepackage{nicefrac}       
\usepackage{microtype}      
\usepackage{lipsum}
\usepackage{fancyhdr}       
\usepackage{graphicx}       
\usepackage{amsmath}
\usepackage{bbm} 
\usepackage{url}
\usepackage{makecell}
\graphicspath{{media/}}     
\usepackage{natbib}
\usepackage{xcolor}
\usepackage{dsfont}
\usepackage{tabularx}

\usepackage{amsthm}

\newtheorem{definition}{Definition}

\usepackage{pgffor}
\usepackage{amsthm}

\usepackage{booktabs}
\usepackage{array}

\newcommand{\Er}[2]{\mathbb{E}_{#2}\left[#1\right]}

\foreach \x in {A,...,Z}{
\expandafter\xdef\csname 
\x c%
\endcsname{\noexpand\ensuremath{\noexpand\mathcal{\x}}}
\expandafter\xdef\csname 
\x til%
\endcsname{\noexpand\ensuremath{\noexpand\tilde{\x}}}
\expandafter\xdef\csname 
\x bb%
\endcsname{\noexpand\ensuremath{\noexpand\mathbb{\x}}}
\expandafter\xdef\csname 
\x b%
\endcsname{\noexpand\ensuremath{\noexpand\mathbf{\x}}}
\expandafter\xdef\csname 
\x h%
\endcsname{\noexpand\ensuremath{\noexpand\hat{\x}}}
\expandafter\xdef\csname 
\x wh%
\endcsname{\noexpand\ensuremath{\noexpand\widehat{\x}}}
}

\foreach \x in {a,...,z}{
\expandafter\xdef\csname 
\x til%
\endcsname{\noexpand\ensuremath{\noexpand\tilde{\x}}}
\expandafter\xdef\csname 
\x b%
\endcsname{\noexpand\ensuremath{\noexpand\mathbf{\x}}}
\expandafter\xdef\csname 
\x h%
\endcsname{\noexpand\ensuremath{\noexpand\hat{\x}}}
\expandafter\xdef\csname 
\x wh%
\endcsname{\noexpand\ensuremath{\noexpand\widehat{\x}}}
}

\edef \greek {alpha, beta, gamma, Gamma, delta, Delta, eps, ve, zeta, eta, theta, Theta, vt, iota, kappa, lambda, Lambda, mu, nu, xi, Xi, pi, Pi, rho, vr, sigma, Sigma, tau, upsilon, Upsilon, phi, vp, Phi, chi, psi, Psi, omega, Omega}

\foreach \x in \greek {
\expandafter\xdef\csname 
\x h%
\endcsname{\noexpand\ensuremath{\noexpand\hat{ {\csname \x \endcsname}}}}
\expandafter\xdef\csname 
\x wh%
\endcsname{\noexpand\ensuremath{\noexpand\widehat{ {\csname \x \endcsname}}}}
\expandafter\xdef\csname 
\x til%
\endcsname{\noexpand\ensuremath{\noexpand\tilde{ {\csname \x \endcsname}}}}
}

\usepackage{multirow}
\title{Unifying Corroborative and Contributive Attributions in Large Language Models}

\author{\textbf{Theodora Worledge}$^*$\qquad \textbf{Judy Hanwen Shen}$^*$\qquad \textbf{Nicole Meister}\qquad \textbf{Caleb Winston}\qquad \textbf{Carlos Guestrin}$^{1,2}$\\ \\ Stanford University}

\begin{document}
\maketitle
\def\thefootnote{*}\footnotetext{Equal contribution. Correspondence to: \{worledge, jhshen\}@stanford.edu}
\def\thefootnote{1}\footnotetext{Chan Zuckerberg Biohub}
\def\thefootnote{2}\footnotetext{Stanford Institute for Human-Centered Artificial Intelligence (HAI)}

\begin{abstract}
As businesses, products, and services spring up around large language models, the trustworthiness of these models hinges on the verifiability of their outputs. However, methods for explaining language model outputs largely fall across two distinct fields of study which both use the term "attribution" to refer to entirely separate techniques: citation generation and training data attribution. In many modern applications, such as legal document generation and medical question answering, both types of attributions are important. In this work, we argue for and present a unified framework of large language model attributions. We show how existing methods of different types of attribution fall under the unified framework. We also use the framework to discuss real-world use cases where one or both types of attributions are required. We believe that this unified framework will guide the use case driven development of systems that leverage both types of attribution, as well as the standardization of their evaluation.

\end{abstract}

\section{Introduction}

The rapid rise of large language models (LLMs) has been accompanied by a plethora of concerns surrounding the trustworthiness and safety of the LLM outputs. For example, these models can ``hallucinate" or fabricate information in response to straightforward prompts \citep{azamfirei2023large}. Beyond simply verifying that generated content can be trusted, knowing the source from which the output was generated is also crucial in many applications. In fact, Bommasani et. al.~\citep{bommasani2021opportunities} highlight that ``\textit{Source tracing is vital for attributing ethical and legal responsibility for experienced harm, though attribution will require novel technical research}". The ubiquitous usage of LLMs in applied settings motivates the development of explanations that provide \textit{both} sources that verify the model output and training sources that are influential in the generation of the output. 
Unfortunately, attributing an LLM output to sources has been mostly studied in two disjoint fields: citation generation and training data attribution (TDA). Verifying the correctness of model outputs, generally situated in the natural language processing community, includes several different tasks such as fact-checking~\citep{yue2023automatic}, knowledge retrieval~\citep{guu2020retrieval, gao2023enabling}, attributed question answering ~\citep{bohnet2022attributed}, and verifiability in language generation \citep{rashkin2021measuring}. Training data attribution, generally situated in the core machine learning community, encompasses a variety of techniques to explain model behavior such as influence functions \citep{koh2017understanding},  data simulators \citep{guu2023simfluence}, and data models \citep{ilyas2022datamodels}. Meanwhile, the term ``\textit{attributions}" is used in both fields. When contemplating the two types of attributions, we can think of the former as external validity, which verifies that the output is correct according to external knowledge, and the latter as a certification of internal validity, which provides the source of the generated content. We can easily imagine applications where both types of validity are important for understanding LLM outputs. For instance, a potential criteria to use for identifying a case of model memorization is for a training source to exactly match the model output while also being highly influential in the generation of the output.

In this work, we argue for a unifying perspective of the citation generation and TDA forms of attribution, which we call \textbf{corroborative} and \textbf{contributive} attributions, respectively. We precisely define each type of attribution and discuss different properties that are desirable in different scenarios. 
Our work provides a first step towards a flexible, but well-defined notion of language attributions to encourage the development and evaluation of attribution systems capable of providing rich attributions of both types.

\subsection{Our Contributions}

\begin{enumerate}
    \item We present an interaction model for LLM attributions that unifies corroborative and contributive attributions through their common components (Section \ref{sec:model}). 
    \item To complete our unified framework, we outline properties relevant to both types of attributions (Section \ref{sec:properties}).
    \item We discuss existing implementations of corroborative and contributive attributions (Section \ref{sec:curr_methods}).
    \item We outline scenarios where attributions are important and discuss their desired properties (Sections \ref{sec:applications}, \ref{sec:case studies}).
    \item We provide directions for future work on attributions (Section \ref{sec:future work}). 
\end{enumerate}


\section{Motivation: The Necessity of a Unified Perspective}

We argue for the study of LLM attributions through a unified perspective of corroborative and contributive attributions. First, we describe the limitations of the current fragmented approach to attributions and then we summarize the case for unification. 
\subsection{Gaps in existing approach to language model attributions}


\paragraph{Misalignment between TDA methods and their use cases} Most training data attribution (TDA) papers present their methods as standalone solutions for motivating use cases such as identifying mislabeled data points \citep{koh2017understanding, yeh2018representer, pruthi2020estimating, schioppa2021scaling, kwon2023datainf}, debugging domain mismatch \citep{koh2017understanding}, and understanding model behavior \citep{grosse2023studying}. In the setting of language models, however, TDA methods may not be a comprehensive solution; training sources that are irrelevant to the content of the test example may be flagged as influential by TDA methods \citep{grosse2023studying}. This is undesirable because the semantic meaning of a flagged training source can indicate its importance in generating the semantic meaning of the output. For instance, when searching for misleading training sources in a Question Answering (QA) language model, it is important to understand which of the sources flagged by TDA methods corroborate the misinformation in the output. This is also the case in other practical applications, such as debugging toxicity. Without carefully considering the types of attribution needed in different use cases, we risk investing in methods that, while establishing essential foundations, may not align with practical use. 
\paragraph{Citation generation methods do not explain model behavior}

Corroborative methods (e.g., fact checking~\citep{yue2023automatic}, citation generation~\citep{guu2020retrieval}) are not designed to explain model behavior. For example, the verifying the truthfulness of outputted facts using sources from an external corpus does little to explain why the model generated such an output. When outputted facts are found to be incorrect, there is limited recourse for correcting model behavior. Thus, corroborative attributions alone cannot address all the challenges of explaining the outputs of language models.

\paragraph{Emergent usage of language models require a richer notion of attributions} 
The emerging use of LLMs in domains such as health care and law involves tasks such as document generation and domain-specific QA that require both explanations of whether the output is correct and where the output came from. As an example, in the legal domain, different products based on LLMs such as legal QA, immigration case document generation, and document summarization are currently under development.\def\thefootnote{\arabic{footnote}}\footnote{Y-Combinator companies in this area include Casehopper, Lexiter.ai, DocSum.ai, and Atla AI.} In this setting, \emph{corroborative} attributions are important to ensure that a generated legal document follows local laws. The sources for such corroborative attributions need not be in the training data. Simultaneously, \emph{contributive} attributions are important for understanding the training documents from which the generated legal document is borrowing concepts. In the legal setting, context and subtle changes in wording matter \citep{bommasani2021opportunities}. 

\subsection{Motivating a unified framework of attributions}

Developing a standardized language to describe different types of attribution will improve the (1) \textbf{clarity} and (2) \textbf{simplicity} of scholarly discussion around attributions. Furthermore, identifying the common components of all attributions provides (3) \textbf{modularity} for improving individual components and better (4) \textbf{reproducibility} of results. Looking ahead to future work, a unified perspective motivates the (5) \textbf{hybrid development} of both corroborative and contributive attributions. 


\paragraph{"Attribution" is an overloaded, ambiguous term} The term "attribution" is overloaded in machine learning literature. Moreover, recent works have attempted to provide both types of attribution for language models under the vague umbrella term of ``attributions'' \citep{bohnet2022attributed, park2023trak, grosse2023studying}. While existing work recognizes the importance of both corroborative and contributive attribution \citep{huang2023citation}, comparing these two notions is difficult without precisely delineating between them while also acknowledging their similarities. A unified perspective of both types of attributions improves the \textbf{clarity} of technical progress on attributions. 

\paragraph{Attribution methods exist concurrently in disjoint fields} The two dominant interpretations of attributions for language model outputs come from the natural language processing (NLP) and explainability communities. In NLP literature, attributing a model output to a source generally refers to identifying a source that corroborates the output \citep{rashkin2021measuring, bohnet2022attributed, yue2023automatic, liu2023evaluating}. We refer to this as \emph{corroborative attribution}. This differs from TDA work, where attributing a model output to a source refers to identifying a training source that highly influenced the model to produce that output \citep{park2023trak, guu2023simfluence, lundberg2017unified, koh2017understanding}. We refer to this as \emph{contributive attribution}. To the best of our knowledge, there is no established framework that unifies these different types of attributions. Furthermore, methods to achieve both types of attribution and metrics to evaluate them have been developed separately. Our goal is to introduce \textbf{simplicity} in understanding the vast landscape of prior work by creating a shared language to discuss attribution methods across different tasks.

\paragraph{Attributions have common components} Despite these two types of attribution being studied in different fields, there are commonalities in system components, properties, metrics, and evaluation datasets. For example, fact-checking using corroborative attributions has significant overlap with fact-tracing using contributive attributions, in terms of metrics and evaluation datasets \citep{akyurek2022towards}. Defining the shared components of different types of attributions introduces \textbf{modularity} that better enables the improvement of individual components of attribution systems. Furthermore, precise definitions of properties shared across different attributions allow for better \textbf{reproducibility} in implementations of attribution systems.

\paragraph{A unifying perspective enables the development of richer attribution systems} Because both notions of attribution are relevant to use cases that improve the safety and reliability of language models as information providers, both are often simultaneously relevant in application settings. There are real-world use cases of attribution that require careful reasoning and differentiating between these two interpretations; some use cases even require both notions of attribution. These use cases should motivate the \textbf{hybrid development} of methods that provide both citation and TDA for LLM outputs. Furthermore, methods used in one type of attribution may be leveraged to develop other types of attributions.

\section{Related Work}
\label{sec:relatedwork}

The majority of prior work has focused on corroborative and contributive attributions separately. Works that have considered both types of attribution in the same setting often do so for specific case studies or experiments without attempting to provide a conceptual unification. This section discusses existing attribution frameworks, as well as works that simultaneously employ notions of corroborative and contributive attributions.



\paragraph{Corroborative attribution frameworks}

Previous work has proposed and leveraged frameworks for attributions that identify supporting sources for model outputs. Notably, \citep{rashkin2021measuring} define a specific notion of corroborative attribution; a model output is transformed into an interpretable standalone proposition $s$, which is then attributed to a source $P$ if it passes the human intuitive test that "According to $P$, $s$". Their \textit{attributable to identified sources} (AIS) evaluation framework evaluates both steps of this definition with human annotators who first evaluate the interpretability of the model output and then whether it satisfies the aforementioned intuitive test for a particular source. Bohnet et. al.~\citep{bohnet2022attributed} applies the AIS framework to the QA setting. Gao et. al.~\citep{gao2023enabling} extends the AIS framework to evaluating LLMs that output citations alongside standard text generations. Another line of work focuses on building and using automated AIS evaluations \citep{honovich2022true, gao2023rarr}. In contrast to prior work, we generalize the definition of corroborative attribution beyond the notion of an "intuitive test" and construct a framework to unify these attributions with contributive attributions.

\paragraph{Contributive attribution frameworks}
Existing TDA work has revealed a common framework for contributive attributions. This shared framework, explicitly defined as data attribution in \citep{park2023trak}, specifies that given a model, list of training data instances, and input, a data attribution method is a function that produces a scalar score for each training instance, indicating the importance of that training instance to the model's output generated from the input. Several lines of work fit under this framework, including influence functions, which make great efforts to scale implementations in the face of significant computational requirements \citep{koh2017understanding, schioppa2021scaling, park2023trak, grosse2023studying, kwon2023datainf}. Surveys summarizing this area include broad categorizations across gradient-based and retraining-based methods~\citep{hammoudeh2022training} and language-specific summaries~\citep{madsen2022post}. 

\paragraph{Shared settings for corroborative and contributive attributions} Even without a shared framework, attributions that are simultaneously corroborative and contributive have naturally appeared. The first of these settings is \textit{fact tracing} \citep{akyurek2022towards}, which recovers the training sources that cause a language model to generate a particular fact. \citep{akyurek2022towards} propose FTRACE-TREx, a dataset and evaluation framework with the explicit goal of identifying corroborative training sources using contributive attribution methods. \citep{park2023trak} also uses FTRACE-TREx as a benchmark for different TDA methods. Another shared setting of corroborative and contributive attributions is the TF-IDF filtering employed in \citep{grosse2023studying}. Here, TF-IDF scores \citep{ramos2003using} are used to filter the training data to a manageable number of sources for influence estimation. While the ultimate objective of this heuristic in \citep{grosse2023studying} is to overcome the bottleneck of training source gradient calculations, the TF-IDF filtering ensures that all of the sources examined are semantically related, which we consider a corroborative notion, to the model input. As the models and training dataset sizes of LLMs continue growing larger, filtering strategies built on notions of corroboration may become the norm. Lastly, \citep{huang2023citation} discuss attributions to non-parametric content, meaning corroborative sources, and attributions to parametric content, meaning contributive sources. While it is perhaps the closest existing work to ours in that it makes explicit the value of both corroborative and contributive attributions, \citep{huang2023citation} largely focuses on roadblocks to practical implementations and pitfalls of attributions in LLMs; a formal unifying framework for the different types of attribution is not proposed.
\section{Formal Problem Statement} 
\label{sec:model}

\begin{figure*}
    \centering
    \includegraphics[width=\textwidth]{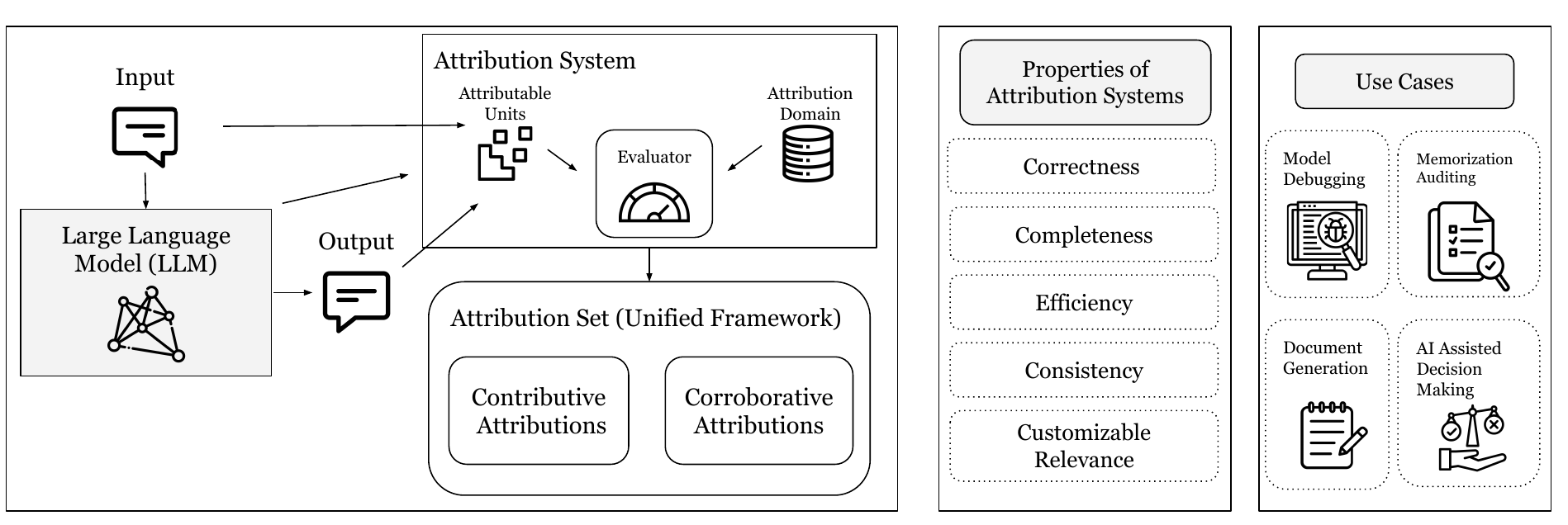}
    \caption{Overview of our proposed unified framework for large language model attributions. We include tasks that require both contributive and corroborative attributions and properties that apply to both types of attributions.}
    \label{fig:overviewfig}
\end{figure*}
\subsection{Interaction Model}
To frame our discussion of attributions for LLMs, we first define the relevant components of an attribution. We build upon the \textit{Attributable to Identified Sources} definition introduced by Rashkin et. al. \citep{rashkin2021measuring} to introduce a general framework for different types of attributions. We define 6 high-level components of the attribution system interaction: the input, model, output, attributable unit, attribution domain and evaluator that allow us to construct an attribution set. As a running example throughout the paper, we consider the use case of attributions for QA in which 
a model provides a short-form output for a given input.



\paragraph{Input} The input is the query provided to the model ($x$). Following the requirements for input interpretability proposed in \citep{rashkin2021measuring}, we assume that $x$ contains the wall-clock time at which it was used to query the model. We consider a variety of different input queries including knowledge queries and generative queries. Knowledge queries are questions that can be answered with the correct piece of information; this is analogous to the QA task. Our scope includes both \textit{Open-book QA} and \textit{Closed-book QA} \citep{roberts-etal-2020-much}. 
Generative queries may have many different answers but may nevertheless require attribution. For example: ``Plan a fun weekend in San Francisco" and ``Write me a Python program to approximate pi'' are both generative queries that require verification before a model can be trusted. While we do not directly consider other interactive settings where there are multiple inputs (e.g., information-seeking dialog \citep{nakamura-etal-2022-hybridialogue} and social-oriented settings (e.g., chit-chat dialog), these are important future directions in which our framework for attribution should extend. \textit{Example input: What is the diameter of the moon?}

\paragraph{Output} The output ($y$) is the response of a language model to the input ($x$). \textit{Example output: 3,475 kilometers} \footnote{https://nightsky.jpl.nasa.gov/club/attachments/Fun\_Facts\_About\_the\_Moon.pdf}

\paragraph{Model} The base language model $M$ takes an input $x$ and generates the output $y$. We note that in practice, some models jointly output attributions with the answer $y$. However, when defining an attribution under our framework, we consider the output generation and attribution generation separately, even if they are generated by the same model. Therefore, for inputs $x \in \mathcal{X}$ and outputs $y \in \mathcal{Y}$, we define \textit{the model} as $M: \mathcal{X} \rightarrow \mathcal{Y}$. \textit{Example model: LLM.}

\paragraph{Attributable Unit} 
In some cases, the full output is used to create an attribution. However, in other cases, a sentence may contain many clauses that need to be independently attributed to achieve the desired level of granularity for the attribution. We define an attributable unit $ z = (x, y, i, j)$ where $i$ and $j$ are the beginning and end indices of tokens in $y$ which require attribution. We define the set of all the attributable units as $Z = {z_1, ..., z_n}$ for $x$ and $y$. \textit{Example attributable set: [("What is the diameter of the moon?", "3,475 kilometers", 0, 15)].}

\paragraph{Attribution Domain} A crucial component of our attribution framework is the domain from which sources (i.e. $s_1, ..., s_m \in D$) for attribution are drawn; we call this the \textit{attribution domain} $D$. There are different promises and limitations when the attributions are drawn from the training data compared to other data not necessarily included in the training. In the practical application and deployment of language models, there are even more domains such as in-context data and fine-tuning data.\footnote{While we leave the complexities of these domains for future work, we discuss in-context data as an attribution domain in Appendix \ref{in context}.} \textit{Example attribution domain: LLM Training Data.}

\paragraph{Evaluator} Each attribution is identified with an evaluation function we call an \emph{evaluator}. Different evaluators lead to different types of attribution. Given an attributable unit $z \in \mathcal{Z}$ and source $s \in \mathcal{D}$, an evaluator $v:\mathcal{Z} \times \mathcal{D} \rightarrow \mathbb{R}$ provides a score that represents the extent to which the given source is an attribution for the attributative unit. In some cases, this value is binary and in others it is continuous. For instance, exact match (EM) is an example of a binary evaluator, which is defined as: 
\[
v_{\text{EM}}(z, s) = \begin{cases}
1 \quad \text{If $y[i:j]$ exists word-for-word within $s$},\\
0 \quad \text{otherwise}.
\end{cases}
\]

Implementations of $v$ are denoted as $\hat{v}$.  An implemented evaluator $\hat{v}$ is not infallible, making it important to evaluate the evaluator against other evaluators on common ground, i.e., potentially using another implementation of the evaluator to compute relevant metrics (see Section \ref{sec:ProperitiesOfAttributionSets}). Past work has used human annotators for $\hat{v}$ \citep{liu2023evaluating, menick2022teaching, rashkin2021measuring}, but the high cost in time and resources of human evaluation has motivated model-based implementations of $\hat{v}$ \citep{yue2023automatic}. \textit{Example evaluator: If seeking a corroborative attribution, we can use the textual entailment evaluator, $v_{\text{TE}}$, as defined in Definition \ref{def:corrEval}. If seeking a contributive evaluator, we can use the counterfactual textual entailment evaluator, $v^M_{\text{CTE}}$, as defined in Definition \ref{def:contribEval}}.

\subsection{Attribution Sets}
Having defined the different components of an attribution system, we now present a definition for an attribution. 

\begin{definition}\label{gas}[Attribution Set]
Given an attributable set $Z$, source domain $D$, evaluator $v$, and evaluator cutoff $\alpha \in \mathbbm{R}$, an attribution set $\mathcal{A}$ is the following set of attributions, or pairs of attributable units and sources: 
\[
\mathcal{A}(Z, D, v, \alpha) = \{(z, s) \ | \ z \in Z, s \in D, v(z, s)\geq\alpha\}
\]
\label{def:attrset}
\end{definition}
We present this definition as a class of explanations for language model outputs. The type of attributions provided in the set depends primarily on the evaluator $v$ and attribution domain  $D$. 
Prior work from \citep{rashkin2021measuring} proposes the AIS framework where the evaluator $v$ seeks to satisfy the intuitive test "According to $s$, $z$" for some source $s$ and sentence $z$. Our definition differs from AIS in several ways. Significantly, the evaluator $v$ of our framework is not restricted to the intuitive test and the attributable unit $z$ of our framework is not restricted to sentence-level explicatures. The flexibility of our framework is important in unifying different approaches to attribution.
\subsection{Attribution Sets with Customizable Source Relevance} 

Definition \ref{gas} of an attribution set considers all sources that satisfy the evaluator cutoff for a given attributable unit as equal in value. Sometimes, however, it is important to value certain sources over others, even if all are valid attributions. Different use-cases demand different notions of relevance; among others, the field of information retrieval has studied multiple manifestations of relevance \citep{cosijn2000dimensions}. To accommodate for this, our definition of a relevance function below allows for custom orders of priority among sources.


\begin{definition} \label{RF} [Relevance Function]
Given attributable units $z \in Z$, attribution domain sources $s \in D$, evaluator $v$, and evaluator cutoff $\alpha \in \mathbbm{R}$, a relevance function is defined as $\phi: Z \times D \rightarrow \mathbb{R} \in [0, 1]$ such that if $v(z, s_1)\geq \alpha$, $v(z, s_2)\geq \alpha$, and $\phi(z, s_1) > \phi(z, s_2)$, then $s_1$ is considered to be a better attribution for $z$ than is $s_2$.

\end{definition}

Adding this additional component of source relevance to an attribution set allows for an ordering of sources within the source domain. While this notion of relevance is not integral to an attribution, it is particularly useful for certain applications. We build off of an attribution set to define the following:

\begin{definition}[$r$-Relevant Attribution Set] \label{rrelevant}
Given an attributable set $Z$, source domain $D$, evaluator $v$, evaluator cutoff $\alpha \in \mathbbm{R}$, relevance function $\phi$, and relevance threshold $r \in \mathbb{R}$, an $r$-relevant attribution set $\mathcal{A}$ is the following set of attributions, or pairs of attributable units and sources: 
\begin{align*}
    &\mathcal{A}(Z, D, v, \alpha, \phi, r) = \\ &\{(z, s) \ | \ z \in Z, s \in D, v(z, s)\geq\alpha, \phi(z, s) \geq r\}
    \label{def:rrattrset}
\end{align*}

\end{definition}

Note that the relevance of a source document for an attribution is a function of the attributable unit. Including a relevance threshold in an attribution set is a way to place priority on certain sources within the attribution domain.



\section{Properties of Attributions} \label{sec:properties}
The central question of \textit{why did a language model provide this answer?} can be answered in many different ways. We present two types of attributions that correspond to different ways of explaining a model output. Furthermore, we build on existing properties of explanations of LLM outputs to define properties that are relevant to language model attributions.
\subsection{Corroborative and Contributive Attributions} \label{sec:contribcorrob}

\paragraph{Corroborative Attributions}
A vast literature exists around corroborative attributions. Prior works refer to these as \textit{citations} in open-domain QA and retrieval settings \citep{guo2022survey, yue2022c}. 


An attribution set (Definition \ref{def:attrset}) is corroborative if its evaluator is corroborative. Corroborative evaluators compare the information content between an attributable unit and a source drawn from the attribution domain. Formally, we define a corroborative evaluator as follows:





\begin{definition}\textbf{Corroborative Evaluator}.
Let $s \in D$ be a source in the attribution domain and $z=(x, y, i,j) \in Z$ be an attributable unit of the input-output pair. A corroborative evaluator is a binary evaluator such that:
\[
v_{\text{corr}}(z, s) = \begin{cases}
1 \quad \text{If $s$ corroborates $z$,}\\
0 \quad \text{otherwise.}
\end{cases}
\]
Moreover, $v_{\text{corr}}$ is a class of different possible evaluators where "corroborate" can have different meanings. Three common corroborative evaluators are: 
\begin{itemize}
    \item \textbf{Exact Match}: $v_{\text{EM}}$ verifies whether there is an exact match between: $y[i:j]$ and a clause in source $s$. 
    \item \textbf{Valid Paraphrase}: $v_{\text{VP}}$ verifies that $y[i:j]$ written as a declarative sentence in the context of $x, y$ is a valid paraphrase of content in $s$; i.e., the declarative sentence is a rewriting of content in $s$ that preserves its truth conditions. 
    \item \textbf{Textual Entailment}: $v_{\text{TE}}$ verifies that $y[i:j]$, in the context of $x, y$, logically follows from the source $s$.\footnote{We assume $s \in D$ has been chosen such that there are no degenerate cases where $s$ contradicts itself, as this would permit an attribution to any $z$.}

\end{itemize}
\label{def:corrEval}
\end{definition}


The study of linguistics has long recognized the inherent \textit{fuzziness} of natural language and so asserts that logical operations are relaxed to \textit{approximate reasoning} when applied to natural language \citep{zadeh1975concept}. Therefore, the logical operations involved in the valid paraphrase and textual entailment evaluators are actually instances of \textit{approximate reasoning}. In practice, the textual entailment evaluator is either implemented through human reasoning or through automated systems capable of natural language inference (NLI), as discussed further in Section \ref{sec:curr_methods}.

For the valid paraphrase and textual entailment evaluators, the context provided by the original input $x$ and the rest of the output $y \setminus y[i:j]$ may be important. To this end, the spans $y[i:j]$ of each attributable unit can be chosen to correspond to sentence-level \citep{rashkin2021measuring} or clause-level 
\textit{explicatures} (see Appendix \ref{cle}). Rewriting a span as an explicature allows the span $y[i:j]$ to be interpreted in the context of $x$ and $y$. In particular, attributable units corresponding to clause-level explicatures within one sentence of the output allow the sentence to be corroborated through more than one source, rather than requiring a single source to corroborate everything in the sentence. In practice, the attributable set is already predefined in many existing tasks and benchmarks \citep{yue2022c}.

In general, the attribution domain of a corroborative attribution may contain any document regardless of whether it was used to train the model or not. The corroborative attribution set for a model output is independent of the model itself; if another model were to produce the same output, the original corroborative attribution set would still be applicable.

\paragraph{Contributive Attributions}
A contributive attribution set is an attribution set (Definition \ref{def:attrset}) that draws from an attribution domain $D$ that is restricted to training sources and relies upon a contributive evaluator. A contributive evaluator is defined as:

\begin{definition} \textbf{Contributive Evaluator}. \label{def:contribEval}
Let $s \in D$ be a source in the attribution domain and $z = (x, y, i, j)$ be an attributable unit. A contributive evaluator for model $M$ is an evaluator such that:
\[
v_{\text{cont}}^M(z, s) \in [0, 1],
\]
where $v_{\text{cont}}^M(z, s)$ quantifies how important source $s$ is to $M$ (trained on $D$) evaluated on the attributable unit $z$. The counterfactual we compare against is $z$ evaluated on a $M$ trained without $s$ (i.e., trained on $D\setminus s$). 
\begin{itemize}
    \item \textbf{Counterfactual contribution to loss (CCL)}: $v_{\text{CCL}}^M$ quantifies the extent to which the loss on $y$ for input $x$ would be different under the counterfactual model $M_{D\textbackslash s}$, compared to under $M_D$. 
    \item \textbf{Counterfactual contribution to the output (CCO)}:
    Let $y' = M_{D \textbackslash s}(x)$ be the counterfactual output of a model trained without $s$. Then, 
    \[
    v_{\text{CCO}}^M(z, s) = \begin{cases}
    1 \quad \text{If $v_{corr}(z, y') = 0$,}\\
    0 \quad \text{otherwise.}
    \end{cases}
    \]
    Note that $v_{\text{corr}}$ is used to indicate whether $z$ is corroborated by the counterfactual model output, $y'$, rather than by a source. Moreover, $v_{\text{cont}}^M$ is a class of different possible evaluators where "contribute" takes on different meanings with different $v_{\text{corr}}$. Any corroborative evaluator, including those mentioned in Definition \ref{def:corrEval}, can be used to construct a contributive evaluator. We highlight two examples of counterfactual output comparison evaluators: 
    \begin{itemize}
            \item \textbf{Counterfactual Exact Match}: $v_{\text{CEM}}^M$ relies on the corroborative exact match evaluator $v_{EM}$ to indicate whether $y[i:j]$ remains the same or changes, had source $s$ not been present in the training data.
            \item \textbf{Counterfactual Textual Entailment}: $v_{\text{CTE}}^M$ relies on the corroborative textual entailment evaluator $v_{TE}$ to indicate whether claims in $y[i:j]$ in the context of $x$ and $y$ remain the same or change, had source $s$ not been present in the training data. 
    \end{itemize}




\end{itemize}
\end{definition}



We note that the CCL evaluator follows standard machine learning methodology more closely than the CCO evaluator, because it operates on the loss, rather than on the discrete output space of language. Accordingly, prior TDA work implements the CCL evaluator (see Section \ref{currcont}).


A shortcoming of the CCL evaluator is that loss does not convey the semantic content of the output. To address this limitation, we introduce the CCO evaluator.\footnote{The gap between machine learning objectives and practical understanding has been highlighted in other areas. For instance, a critique of mechanistic explanations of model behavior, which solely rely on the inner workings of the model, is that they are not interpretable to humans \citep{erasmus2021interpretability}. Similar to how mechanistic explanations can be enriched by practical explanations that are meaningful to applied practitioners, existing loss based attributions can be enriched by attributions for counterfactual changes in the output semantics.} Keeping with the running example of querying a model with "What is the diameter of the moon?" and it generating the response, "3,475 kilometers", we can imagine using the counterfactual textual entailment CCO evaluator. In this case, a source $s$ would be deemed contributive if its removal from the training set would result in a counterfactual model that outputs "At least 3,000 kilometers" in response to the same input, but not if it outputs "3,475,000 meters". This differs from the CCL evaluator, which identifies a training source as contributive if its removal leads to a counterfactual model that has significantly different loss on the output, regardless of how the semantic meaning of the counterfactual output differs, if at all. We advise that this novel concept of CCO evaluators be a focus of future work on contributive attributions for LLMs.

\subsection{Properties and Metrics of Attribution Sets}
\label{sec:ProperitiesOfAttributionSets}
Depending on the application of the LLM, different properties of attribution sets may be desirable. Crucially, these desiderata may be different from those of general machine learning explanation methods.\footnote{https://christophm.github.io/interpretable-ml-book/properties.html} While properties are high-level qualities that are desirable in an LLM attribution, metrics are specific methods to measure these properties. A single property can be measured by many different metrics. While we provide a few metrics for each property in Table \ref{tab:properties-metrics}, future work may use different metrics for these properties.      

\paragraph{Correctness} The most ubiquitous measure of attribution sets in current work is whether an attribution set is correct. To interrogate properties of correctness, some notion of ground truth, often in the form of an oracle evaluator $v$, is required to properly score each attribution. 
\begin{itemize}
    \item \textbf{Attribution validity:} For each attribution in an attribution set, the notion of validity captures how correct the attribution is relative to a ground truth evaluator. Corroborative attributions generated by various systems have been evaluated for validity using $v_{\text{TE}}$ implemented via human reasoning \citep{rashkin2021measuring, bohnet2022attributed}. Contributive attributions have been evaluated for validity using leave-one-out retraining \citep{koh2017understanding} and the proximal Bregman response function \citep{grosse2023studying}. 
    \item \textbf{Coverage:} An attribution set $\mathcal{A}$ with evaluator cutoff $\alpha$ has perfect coverage if $\forall \ z \in Z \ \exists (z,s) \in \mathcal{A},v(z, s) \geq \alpha$. Previous work has referred to coverage as attribution recall \citep{liu2023evaluating}. One way to measure coverage is to calculate the proportion of attributable units in $Z$ with a valid attribution under an oracle evaluator $v$ included in $\mathcal{A}$ \citep{liu2023evaluating}. 
    \item \textbf{Attribution precision:} Another way to measure attribution set correctness is precision. An implemented attribution set $\hat{\mathcal{A}}$ with evaluator cutoff $\alpha$ is precise if $v(z, s) \geq \alpha \ \forall \ (z, s) \in \hat{\mathcal{A}}$. By definition, an attribution set $\mathcal{A}$ has perfect precision. However, this is an important property when evaluating implementations of attribution systems, where the components analogous to attribution evaluators are imperfect. One way to measure the precision of an attribution set is to calculate the proportion of valid attributions under an oracle evaluator $v$ \citep{liu2023evaluating}.
\end{itemize}


\paragraph{Attribution Recall} Let $S'$ be the set of all documents that provide attribution for a given $z$ (i.e., $S'(z) = \{s | s \in D, v(z, s) \geq \alpha\} $). 
The attribution set $\mathcal{A}$ has perfect recall for $z$ if $\forall s \in S'(z), (z,s) \in \mathcal{A}$. One way to measure the recall of an attribution set for $z$ is to calculate the proportion of sources from the attribution domain that fulfill $v(z, s) \geq \alpha$ that are actually included in the attribution set. This is a measurement of the sources that can attribute one specific $z$, which differs from coverage which focuses on whether all $z \in Z$ is attributed. 

In the corroborative setting, there may be many sources that can provide an attribution for $z$. Attribution recall might be important when an attributable unit $z$ requires multiple sources to validate. For example, facts about the efficacy of certain drugs might require all relevant studies to be included rather than just a single source. In the contributive attribution setting, many training documents may have been influential in generating an output. Having perfect attribution recall is relevant when using attribution to assign credit to training data authors and for model debugging, where all sources need to be identified. Measuring attribution recall has appeared in prior work \citep{pruthi2020estimating} as a measurement of the fraction of artificially mislabeled examples that were successfully identified through gradient tracing for TDA. 

\paragraph{$r$-Relevancy}
As explained in definition \ref{rrelevant}, an attribution set is $r$-relevant if all the sources in the attribution set meet the threshold of $r$ under some relevancy function, $\phi$. $r$-Relevancy is an important property because some applications find certain sources in the attribution domain to be more useful than others. This is the case in the setting of corroborative attributions for fact-checking, where trustworthy sources are more relevant than questionable sources. This is also the case in the setting of corroborative attributions for generating citations for written reports, where primary sources tend to be more relevant than secondary or other derivative sources. Although motivated from an efficiency standpoint, \citep{grosse2023studying} in effect implements $r$-relevant contributive attribution sets with TF-IDF filtering as a relevancy function; only sources that are high in TF-IDF similarity to the input are considered for the attribution set. A metric to measure the $r$-relevancy of an attribution set is the proportion of attributed sources that meet the relevancy threshold $r$.

\subsection{Properties and Metrics of Attribution Systems}
Properties of attribution sets are inherent to a single attribution set. However, some properties are instead functions of the implemented system that generates the attribution sets in the first place. We discuss two such properties.
\begin{table}[]
\center
\begin{tabular}{|l|l|}
\hline
\textbf{Properties}   & \textbf{Metrics} \\ \hline
Correctness  & \begin{tabular}[c]{@{}l@{}} Validity \citep{rashkin2021measuring, bohnet2022attributed}\\ 
Coverage \citep{liu2023evaluating, gao2023rarr} \\ Attribution Precision \citep{liu2023evaluating} \end{tabular} \\ \hline
Attribution Recall & Mislabeled example identification~\citep{pruthi2020estimating}  \\ \hline
Relevancy    & \begin{tabular}[c]{@{}l@{}} Proportion of attribution set that is $r$-relevant \end{tabular} \\ \hline
Consistency/Replicability    & \begin{tabular}[c]{@{}l@{}} Attribution set distance \end{tabular} \\ \hline
Efficiency   & \begin{tabular}[c]{@{}l@{}}Training time \citep{park2023trak, hammoudeh2022training} \\ Inference time \citep{yeh2022first, hammoudeh2022training}\\ Training memory requirements \citep{hammoudeh2022training}\\ Inference memory requirements \citep{hammoudeh2022training}\end{tabular} \\ \hline
\end{tabular}
\caption{Properties of attribution sets and systems. Different metrics have been proposed by prior literature in measuring each of these properties.}
\label{tab:properties-metrics}
\end{table}


\paragraph{Consistency} An attribution system is considered consistent if, for similar inputs and outputs in an attribution domain, the generated attribution sets are similar. For a fixed attributible set $Z$, attribution domain $D$, evaluator $v$, and evaluator cutoff $\alpha$, an attribution system is $\epsilon$-stable over sources of randomness in the system if for $\mathcal{A}$ and $\mathcal{A}'$ sampled from different executions, $\mathbbm{E}[d(\mathcal{A}, \mathcal{A}')] \le \epsilon$, where $d$ is some distance metric defined over input-output pairs and over attribution sets respectively (e.g., $d$ could be the Jaccard distance over sources' indicator functions). 

This property is particularly important when decisions based on LLM outputs need to be documented as justification. For corroborative attributions, a legal service scenario may require documentation of sources for advice provided to customers. For contributive attributions, an authorship compensation scenario would require attribution consistency to fairly determine payments to creators. In both cases, there is value in replicating the same attribution set at a later time with the same inputs.

Prior work highlighting the shortcomings of contributive methods (e.g., influence functions) demonstrates increased variance in influence estimates for deeper models; this would preclude consistency unless influence is estimated using an average across multiple runs~\citep{basu2020influence}. Similarly, averaging gradients across checkpoints during training might lead to inconsistent estimates of influence estimation because the ordering of examples has a significant impact on observed influence~\citep{sogaard2021revisiting}. However, consistency has not been directly measured in prior work for contributive or corroborative attributions.



\paragraph{Efficiency} Efficiency describes the time and space complexity required by an implementation of an attribution system in generating an attribution set for a given attribution domain, input, and output. Prior works on large language models examine both training and inference efficiency in terms of energy cost and CO$_2$ emitted \citep{bender2021dangers, liang2022holistic}. However, attribution systems vary widely in function and implementation. 

In a survey of attribution methods, Hammoudeh et. al.~\citep{hammoudeh2022training} summarize inference time, space, and storage requirements for influence analysis methods as a function of training dataset size, model parameter count, and training iteration count.

\section{Current Methods}
\label{sec:curr_methods}

\subsection{Corroborative Attribution Methods}

Prior work primarily focuses on identifying corroborative attributions with the textual entailment evaluator $v_{\text{TE}}$. Two common approaches to implementing $v_{\text{TE}}$ are human reasoning \citep{rashkin2021measuring, liu2023evaluating} and automated systems capable of natural language inference (NLI) \citep{honovich2022true, gao2023rarr}. Often, NLI systems are used in corroborative attribution systems to identify attributions, whereas human reasoning is used to evaluate attributions and also to generate training data for NLI systems. Both implementations exclude the usage of background information external to the source in judging the entailment relation \cite{honovich2022true}. However, different sets of background knowledge may be leveraged by humans and NLI systems when interpreting the meaning of $s$ and $z$ \citep{rashkin2021measuring}; identifying discrepancies in NLI systems based on background knowledge and human judgment is important for addressing patterns of bias in evaluator performance.

Outside of implementing the evaluator, there are many different design choices to be made when building corroborative attribution systems and it is often unclear which method is the best. This is exacerbated by a lack of standardization in the evaluation metrics and datasets. To demonstrate this, we provide an overview of these implementations in Table~\ref{tab:corrob_methods} and how they align with the interaction model defined in Section ~\ref{sec:model}. 


\begin{table*}
\begin{center}
\begin{tabularx}{\textwidth} { 
  | p{2.5cm}
  | p{1.5cm}
  | p{2cm}
  | p{5.8cm}
  | >{\raggedright\arraybackslash}X
  | >{\raggedright\arraybackslash}X
  | }

  \hline
\textbf{Method} & \textbf{Attributable Unit} & \textbf{Attribution Domain} & \textbf{Model} & \textbf{Evaluator}
\\
\hline
ALCE~\citep{gao2023enabling}
 & Output $y$ parsed into sentences $\{z_1 .. z_n\}$ & Wikipedia (2018-12-20), Sphere~\citep{piktus2022web} (Filtered Common Crawl)
&  
1. \textbf{Retrieval}: Retrieve top 100 passages (using GTR~\citep{ni-etal-2022-large} and  DPR~\citep{karpukhin-etal-2020-dense} for Wikipedia and BM25~\citep{robertson2009probabilistic} for Sphere).

2. \textbf{Synthesis}: Synthesize retrieved passages to identify the $k$ most relevant.

3. \textbf{Generation}: Include these k passages in-context alongside the input and additional prompting that instructs the model to cite the passages used.

& \textbf{Textual Entailment}: NLI model that outputs 1 if the source entails the outputs.
\\ \hline
GopherCITE~\citep{menick2022teaching}
 & Output $y$  & Internet (queried by Google Search)
& 
Collect human preferences of evidence paragraphs that support provided answers. Perform both supervised learning on highly rated samples and reinforcement learning from human preferences on Gopher \citep{rae2022scaling}, to learn a model that finds relevant web pages on the internet and quotes relevant passages to support its response.
  & \textbf{Textual Entailment}: LLM is fine-tuned to perform NLI.
\\ \hline
LaMDA~\citep{thoppilan2022lamda}
 & 
 Output $y$  & Internet (queried by information retrieval system that returns brief text snippets) 
 & 
 Model is fine-tuned to learn to call an external information retrieval system and use the results in-context to generate an attributed output.
 & \textbf{Textual Entailment}: LLM bases its output off of retrieved sources.
\\ \hline
WebGPT~\citep{nakano2021webgpt}
 & Output $y$ parsed into sentences $\{z_1 .. z_n\}$ & Internet (queried by Microsoft Bing Web Search API) & Given a text-based web-browsing environment, GPT-3 is fine-tuned with RLHF to use the browser to identify sources it then uses in-context to answer the query.
 &  
\textbf{Textual Entailment}: LLM bases its output off of retrieved sources.
\\ \hline
Lazaridou et al. (2022) ~\citep{lazaridou2022internet}
 & Output $y$ & Internet (queried by Google Search)
 & 
 1. \textbf{Retrieval}: Extract text from top 20 URLs returned by Google to. 
 
 2. \textbf{Generation}: Use few-shot prompting to steer model to provide an answer conditioning on evidence.
 
 3. \textbf{Attribution}: Rank all the paragraphs from top 20 URLs by cosine similarity between the paragraph and query.
 & \textbf{Cosine similarity} between question and evidence paragraphs.
\\ \hline
RARR ~\citep{gao2023rarr}
 & Output $y$ & Internet (queried by Google Search)
 & 
 1. \textbf{Generation}: For an input, which takes the form of a question, use PaLM \citep{chowdhery2022palm} to generate the output. 
 
 2. \textbf{Retrieval}: Use Google Search to retrieve five web pages and then identify four-sentence evidence snippets from these pages that are relevant to the input, according to GTR \citep{ni-etal-2022-large}.

  3. \textbf{Attribution}: Use chain-of-thought few-shot prompting \citep{wei2023chainofthought} on PaLM \citep{chowdhery2022palm} to identify cases where the evidence snippet and the model output provide the same answer to the input.
 
 & \textbf{Valid Paraphrase:} LLM identifies when the source and model output provide the same answer to the input.
\end{tabularx}
\hrule
\end{center}
\caption{Overview of existing corroborative attribution systems for language models}
\label{tab:corrob_methods}
\end{table*}

In Table~\ref{tab:corrob_eval}, we outline the evaluation metrics used in prior work. Most proposed implementations evaluate attribution outputs with a metric that evaluates the quality of the LLM output, independent of the accompanying attribution, in addition to attribution correctness (Table~\ref{tab:corrob_eval}). To measure the quality of the LLM output, methods often measure the fluency or plausibility of the output to the user. Generally, this involves asking a user if the output is interpretable or helpful, or measuring performance on a QA or classification task (e.g., Exact Match for QA). 
Metrics for measuring correctness of an attribution set assess if the attributed output is fully supported by its corresponding corroborative documents (e.g., attribution precision and coverage).

\begin{table*}
\begin{center}
\begin{tabularx}{\textwidth} { 
  | >{\raggedright\arraybackslash}X 
  | >{\raggedright\arraybackslash}X 
  | >{\raggedright\arraybackslash}X
  | >{\raggedright\arraybackslash}X | }
  \hline
\textbf{Method} & \textbf{Datasets} & \textbf{Non-attribution Evaluation} &\textbf{Attribution Evaluation: Correctness}  \\
\hline
Attributable to Identified Sources (AIS)~\citep{rashkin2021measuring}  &  QReCC and WoW (QA), CNN/DM (summarization), ToTTo dataset (table-to-text task)

 & \textbf{Human Reasoning}: Is all of the information relayed by the system response \textbf{interpretable} to you?
 & \textbf{Human Reasoning}: Is all of the information provided by the system response (a) \textbf{fully supported} by the source document?
\\\hline
 Evaluating Verifiability in Generative Search Engines ~\citep{liu2023evaluating} & AllSouls, davinci-debate, ELI5, WikiHowKeywords, NaturalQuestions (all filtered)
& \textbf{Human Reasoning}: \textbf{Fluency, perceived utility} (whether the response is a helpful and informative answer to the query)
& \textbf{Human Reasoning}: Coverage, citation precision 
\\\hline
Automatic Evaluation of Attribution by Large Language Models ~\citep{yue2023automatic}&
HotpotQA, EntityQuestions, PopQA, TREC, TriviaQA, WebQuestions
& 
\textbf{None} & \textbf{Automatic Evaluation}: Fine-grained citation precision: Is the attribution attributable, extrapolatory, or contradictory?
\\\hline
ALCE~\citep{gao2023enabling} & ASQA, QAMPARI, ELI5 & \textbf{Automatic Evaluation}: Fluency (MAUVE), Correctness (compared to a ground truth answer) measured with exact match and entailment (NLI) & \textbf{Automatic Evaluation}: Coverage, citation precision
\\\hline
GopherCITE ~\citep{menick2022teaching} & NaturalQuestionsFiltered, ELI5Filtered
& \textbf{Human Reasoning}:
Is the answer a \textbf{plausible reply} to the question? & \textbf{Human Reasoning}: Coverage
\\\hline
WebGPT ~\citep{nakano2021webgpt} & ELI5, TruthfulQA
& \textbf{Human Reasoning}: Overall usefulness, coherence
 & \textbf{Human Reasoning}: Factual correctness 
 \\\hline
\end{tabularx}
\end{center}
\caption{Overview of the evaluation of corroborative attributions}
\label{tab:corrob_eval}
\end{table*}


\subsection{Contributive Attribution Methods For Language Models} 
\label{currcont}
Given a model, input, and output, contributive attributions provide a score for each source in the attribution domain that represents the relative amount that the source contributed to the output. The area of TDA for language tasks has been highlighted by Madsen et. al.~\citep{madsen2022post} as a specific interpretability technique. Hammoudeh et. al.~\citep{hammoudeh2022training} give a broader view of different techniques for TDA that are theoretically applicable to language models. However, relatively few works thus far have specifically studied TDA in language models. We broadly categorize the many methods proposed for TDA into two families: data-centric and model-centric techniques. At a high-level, data-centric techniques average the effects of data changes across different models while model-centric techniques interrogate a single model. Since we are concerned with providing attributions for a specific model, we focus on describing verifiers for model-centric techniques. 

\paragraph{Data-Centric TDA}
To understand the impact of data points used to train models, one view is to take averages across different models that are trained without that data point. The common goal of retraining a model with the data point left out (i.e., leave-one-out (LOO) retraining) has been implemented differently by various techniques. 

Let $f \in \mathcal{F}$ where $\mathcal{F}$ is a family of functions parameterized by $\theta$ trained on dataset $D$. Data-centric approaches characterize the influence (e.g., $\mathcal{I}(z_i, z_{te}, D)$) of a data point  $z_i=(x_i, y_i)$ on a test point $z_{te}=(x_{te}, y_{te})$ over dataset $D$ as an average effect over many possible models. For instance, LOO influence is the following:  
\begin{align}
    &\mathcal{I}_{LOO}(z_i, z_{te}, D) =  \nonumber\\&\Er{L(f(x_{te}, \theta_{D\setminus z_i}), y_{te} ) - L(f(x_{te}, \theta_{D}), y_{te})}{f \in \mathcal{F}}. \label{eq:data centric influence}
\end{align}

\noindent For LOO retraining, the effect of leaving one example out is averaged over different training runs removing the effect of the randomness of training. Approximations to LOO such as Datamodels \citep{ilyas2022datamodels} compute an average across leaving different subsets of points out and use the difference between logits as the functions $L$. Data Shapley Values \citep{ghorbani2019data} approximates this expectation using different possible subsets of the entire dataset. For Data Shapley, we can think of $\mathcal{F}$ as the family of functions induced by different subsets $D' \in D\setminus z_i$: 
\begin{align*}
    & \mathcal{I}_{DS}(z_i, z_{te}, D) \\ = &\frac{1}{n} \sum_{D' \in D\setminus {z_i}} \frac{1}{{n-1 \choose | D'|}}L(f(x_{te}, \theta_{D'}), y_{te} ) - L(f(x_{te}, \theta_{D' \cup {z_i}}), y_{te}).
\end{align*}s
These methods explicitly compute, approximate, or learn to predict counterfactual changes to the loss with one example removed.  

\paragraph{Model-Centric TDA}
For methods that aim at understanding and attributing a specific model, only parameters for a single model or a single training trajectory are considered. The \emph{counterfactual contribution to loss} evaluator ($v^M_{\text{CCL}}$) is an abstraction of the notion of attribution in this section. Methods in this area take the following general form: 
\begin{align}
    &\mathcal{I}_{MC}(z_i, z_{te}, D) = \nonumber\\& \mathbbm{E}_{f \in \mathcal{F}}[L(f(x_{te}, \theta_{D\setminus z_i}), y_{te})] - L(f(x_{te}, \theta_{D}), y_{te}).
    \label{eq:model centric influence}
\end{align}

\noindent While Equation (\ref{eq:data centric influence}) takes an expectation of both terms over $\mathcal{F}$ parameterized by $\theta$ trained on dataset $D$, Equation (\ref{eq:model centric influence}) only takes this expectation over the counterfactual term that excludes $z_i$ from training. Therefore, $\mathcal{I}_{MC}(z_i, z_{te}, D)$ is relative to a specific model's loss, rather than to an expected model's loss. 

Influence functions~\citep{koh2017understanding, bae2022if} fall within this category because they approximate the expectation in the first term of $\mathcal{I}_{MC}(z_i, z_{te}, D)$ by modeling the response induced by upweighting $z_i$ on model $\theta_{D}$. Influence function methods estimate the counterfactual effect of individual training examples on model predictions for an individual model \citep{koh2017understanding, park2023trak, kwon2023datainf}. Further work acknowledges that when applied to nonconvex learning objectives, influence functions more closely estimate the Proximal Bregman Response Function, rather than the counterfactual influence \citep{bae2022if, grosse2023studying}. All of these methods are implementations, even if computationally impractical for today's LLMs, of the counterfactual contribution to loss $v_{\text{CCL}}^M$ evaluator. 



For Gradient Tracing methods, such as TracIn \citep{pruthi2020estimating}, the quantity measured is different from all the definitions above and we believe it lacks the explicit counterfactual motivation needed for contributive attributions. Specifically, the ideal objective function of TracIn seeks to measure the contribution of an example to the loss over the training process by summing the change in loss across training time steps that include $z_i$ in the batch:

\small
\begin{align*}
      \mathcal{I}_{TI}(z_i, z_{te}, D) &= \sum_{t: z_i \in B_t} L(f(x_{te}, \theta_{t-1}), y_{te}) - L(f(x_{te}, \theta_t), y_{te}). 
\end{align*}
\normalsize
TracIn does not explicitly define a relationship between its notion of influence of a training point $z_i$ and the final model's behavior on the test point $z_{te}$. Therefore, this method does not fall within our framework of counterfactual evaluators. 


\begin{table*}[]
    \centering
    \begin{tabular}{|c|p{4cm}|p{4cm}|c|}
    \hline 

      \textbf{Method Type}     & \textbf{Oracle Evaluator} & \textbf{Implemented Evaluator} & \textbf{LM Implementations} \\ \hline
       \multicolumn{4}{|c|}{\textbf{Data-Centric Methods}} \\ \hline    
      Leave-one-out       & Change in the expected  counterfactual output & Expected counterfactual contribution to the loss & DataModels~\citep{ilyas2022datamodels}   \\ \hline 
      Shapley Values            & Change in the expected counterfactual output  & Expected counterfactual contribution to the loss & Data Shapley \citep{pmlr-v97-ghorbani19c}   \\ \hline
      \multicolumn{4}{|c|}{\textbf{Model-Centric Methods}} \\ \hline
      Influence Functions  & Change in the counterfactual output ($v^M_{\text{CCO}}$) & Counterfactual contribution to the loss ($v^M_{CCL}$) & \makecell{TRAK \citep{park2023trak}\\ EK-FAC \citep{grosse2023studying}}\\ \hline
      Gradient Tracing & Change in training trajectory & Contribution to the loss &  \makecell{TracIn \citep{pruthi2020estimating}  \\ Simfluence \citep{guu2023simfluence} \\ TracIN-WE \citep{yeh2022first} }  \\ \hline 
    \end{tabular}
    \caption{Overview of contributive attribution methods for language models}
    \label{tab:my_label}
\end{table*}



\section{Use Cases Requiring Attributions}
\label{sec:applications}

\begin{table}[]
    \centering
    \begin{tabular}{|p{4.4cm}|c|c|c|c|c|}
    \hline 
      \multirow{2}*{\textbf{Task}} & \multicolumn{5}{c|}{\textbf{Properties}} \\ \cline{2-6}
       & Correct. & High Recall & Effici. & Consist. & Relev. \\ \hline
      \multicolumn{6}{|c|}{\textbf{Corroborative Attribution}} \\ \hline
      Question Answering & \checkmark & & \checkmark & & \checkmark \\ \hline
      Fact Checking & \checkmark & & \checkmark & & \\ \hline
      \multicolumn{6}{|c|}{\textbf{Contributive Attribution}} \\ \hline
      Author Compensation & \checkmark & \checkmark & \checkmark & \checkmark & \\ \hline
      GDPR Compliance & \checkmark & \checkmark & \checkmark & \checkmark & \\ \hline
      Model Bias Detection & \checkmark & \checkmark & \checkmark & & \\ \hline
      \multicolumn{6}{|c|}{\textbf{Contributive+Corroborative Attribution}} \\ \hline
      Model Debugging & \checkmark & \checkmark & \checkmark &  & \\ \hline
      Auditing Model Memorization & \checkmark & \checkmark & \checkmark & \checkmark &  \\ \hline
      Human AI Collaboration & \checkmark & \checkmark & \checkmark & \checkmark & \checkmark \\ \hline
    \end{tabular}
    \caption{Overview of attribution use cases and their desired properties.}
    \label{tab:use_cases}
\end{table}




While perhaps the most obvious use case of attributions is to provide citations for
a model's answer to a question, the interaction model we have presented obviates
a number of use cases, each with its own list of desirable properties. Across the board, the properties of correctness and high efficiency are important.
Depending on the use case, either contributive attributions, corroborative attributions, or a composition of the two are required.
In this section, we enumerate use cases and our recommendation on how to apply attributions. 



\subsection{Use Cases of Corroborative Attributions}

While there are a variety of use cases where corroborative attributions are important, we highlight several tasks that showcase how different attribution properties and metrics are meaningful.

\paragraph{Question Answering} QA is a common task for LLMs.
Unfortunately, LLM answers are not always trustworthy, especially in critical domains such as law and healthcare \citep{choudhury2023investigating}. 
\citep{bohnet2022attributed} and QA engines such as Bing Chat and Perplexity AI have explored using corroborative attributions to provide citations for answers \citep{khan2023unstoppable}. In this use case, humans can verify the output by examining the sources that are provided as attributions. This step of output verification by the human user is critical because the attribution domain may not be fully composed of trusted sources (e.g., QA engines retrieve from the internet).

High attribution recall is not a strict requirement for QA since only a few corroborating sources may be sufficient to support an attributable unit. Implementations of attribution for QA may customize source relevance to prioritize primary sources, rather than secondary sources, or more reputable sources, rather than those from authors of dubious credentials.





\paragraph{Fact Checking}
Fact checking has emerged as a promising tool in the fight against
misinformation~\citep{krause2020fact}.
Despite its importance, fact checking has long been an entirely manual 
process~\citep{amazeen2015revisiting}.
Many researchers have attempted to automate fact checking~\citep{guo2022survey}.
We posit that our attributions framework can help create and evaluate
methods for fact checking.

Given an attribution domain of sources that are up-to-date, trustworthy, and non-contradictory, it follows that an attributable unit can be taken as true if it has at least one corroborative attribution. Therefore, high attribution recall is not an important property for this use case. As in the QA use case, customized source relevance can be useful for prioritizing primary sources. However, because the attribution domain is assumed to contain only trustworthy sources, customized source relevance is redundant to the end of selecting trusted sources.

Interestingly, perfect coverage is \textit{not} necessarily desired in this use case; low coverage indicates that either the output is nonfactual or that the attribution domain does not include sufficient sources to corroborate the statement. If the model output is factual, however, the coverage should be perfect. Coverage is perhaps a numerical counterpart to non-binary labels for factuality, such as "mostly true" or "half true", from previous work \citep{guo2022survey}.

This setting of fact-checking motivates another class of corroborative evaluators that indicates a lack of logical entailment. For example, an evaluator that indicates when a source contradicts an attribution unit would make it possible to flag a model output for containing misinformation. Prior work has implemented such evaluators before; RARR \citep{gao2023rarr} first identifies sources that are relevant to an LLM output and then post-edits unattributed parts of the output by using an agreement model that indicates when part of the output disagrees with a source.




\subsection{Use Cases of Contributive Attributions}

Prior work has explored using contributive attributions to understand
the training data of models. We discuss some of these tasks and their desired properties here. 

\paragraph{Author Compensation}
With LLMs being trained on large datasets that include sources
under various licenses, people have begun to observe language models
returning output that heavily resembles licensed works owned by specific authors.
As a result, thousands of authors have demanded compensation for
their work being used to train language models~\citep{samuelson2023generative}.
This demand necessitates the ability to attribute language model output to
specific author sources and to quantify the degree to which the author's work
contributed to the output.

In this use case, authors could be compensated based on their work appearing in the contributive attributions of an LLM output.
High attribution recall and consistency are critical since leaving out a major contributor could have legal consequences.

\paragraph{GDPR Compliance}
GDPR compliance requires language model maintainers to update their models
by removing the influence of training data upon request. Prior work has explored efficient data
deletion for ML models~\citep{brophy2020exit} to avoid training from scratch
with a few data points removed. In such a scenario, it is critical to ensure
that the original data points are no longer contributing to the model output.

An empty contributive attribution set for a
set of language model outputs can imply the deleted data is no longer
influential.
The attribution set must have high attribution recall or else an empty set may be a false positive
for compliance. For the same reason, stability is also critical.
\subsection{Use Cases 
 of Corroborative \textit{and} Contributive Attributions}
\label{corr_and_contr_use_cases}
We describe several use cases that require both corroborative and contributive attributions for LLM predictions. 

\paragraph{Model Debugging} Identifying the training data points that contribute to a test case that is incorrect, or otherwise undesirable (e.g., toxic), is helpful for cleaning the training data and remedying the failure case in the model development cycle.\footnote{Retraining an LLM from scratch is too resource intensive to be practical. However, the fine-tuning process is less resource intensive and more reasonable to repeat; attributions for fine-tuned model outputs to fine-tuning data may be the most actionable setting for debugging models with attributions.} While this has been a longstanding motivation of TDA papers \citep{koh2017understanding, yeh2018representer, pruthi2020estimating, schioppa2021scaling}, we argue that when working with language models, not only do we need contributive attributions, but we also need corroborative notions of attribution. This is because TDA methods are not guaranteed to flag training sources that are semantically relevant to the input and output \citep{grosse2023studying}; removing semantically \textit{unrelated} contributive sources is not guaranteed to change the semantic meaning of the model output. Therefore, the semantic relation between contributive sources and the input and output is important for model debugging. Corroborative attributions are integral in identifying such semantic relation. \textit{Data poisoning detection} \citep{koh2017understanding} is adjacent to model debugging and thus requires the same types of attribution. 




\paragraph{Document Generation} When given a prompt, the drafting task describes the language model of writing a passage of text. A growing number of ventures are now proposing using LLMs for writing documents such as legal briefs and contracts (Section \ref{sec:legal}). In this task, both types of attributions are helpful for the generated output $y$. Contributive attributions would provide context for what sources the generated documents are similar to and corroborative attributions would provide validation for the claims made in the generated document. 

\paragraph{Auditing Model Memorization} To determine that an output is a case of model memorization of a training point, the output must exactly match a training point that was also highly influential in its generation. Therefore, this use case requires exact match corroborative attributions, as well as contributive attributions. Prior work has measured the extent to which models have memorized their training sources via self-influence, defined as the influence of a training point on its own loss \citep{feldman2020neural, pruthi2020estimating}. However, this approach does not extend to the evaluation of inputs from outside the training set. Furthermore, we believe that heuristic approaches that solely use corroborative exact match to diagnose cases of model memorization exclude contributive attributions due to the inefficiency of current TDA methods. 

\paragraph{Human-AI Collaboration}
Another rapidly emerging use case is using LLMs for human-AI collaboration. For example, Sun et. al.~\citep{sun2022investigating} study AI-supported software engineering through several language model collaborative tasks. In their study, participants wanted to know how the code was generated (i.e., contributive attribution) as well as code correctness (e.g., corroborative attributions). Liao et. al.~\citep{liao2020questioning} summarize a broader family of AI-assisted tasks such as including decision support and communication support; study participants wanted to know what training data produced the model suggestion as well as the correctness of the suggestion. Furthermore, in application domains such as assistive call center tools or travel itinerary tools, companies are using LLMs for various collaborative planning and decision tasks.\footnote{Start-ups in this area include Observe AI, GenixGPT} In Human-AI collaboration tasks, all of the properties we describe may be important. Particularly, when a task process is documented, consistency in the attribution provided for making such a decision is important. In this example, both types of attributions are desired for the same output $y$ of a language model. 

\section{Case Studies: A Closer Look at Two Application Domains}
\label{sec:case studies}
\subsection{Case Study 1: LLMs for Legal Drafting}
\label{sec:legal}
AI and LLMs in particular have been increasingly applied to the legal domain as training data for different legal tasks are becoming more readily available \citep{hendersonkrass2022pileoflaw, guha2023legalbench,niklaus2023multilegalpile,cui2023chatlaw,shaghaghian2021customizing,nay2023large}. While LLMs show promising results for legal document analysis, contract review, and legal research, they also raise concerns regarding privacy, bias, and explainability~\citep{sun2023short,deroy2023ready}. To address such shortcomings, the development of attribution methods to promote transparency and interpretability are needed~\citep{sun2023short}.
Moreover, Bommasani et. al.~\citep{bommasani2021opportunities} discuss the opportunities and risks of using foundation models for applications rooted in US law in particular. They review different fields of law and specifically contemplate the ability of foundation models to write legal briefs. While tools for writing legal briefs using language models are still under development, different products based on LLMs such as legal question answering, immigration case drafting, and document summarization have started to appear in various startups.\footnote{Y-Combinator companies in this area include Casehopper, Lexiter.ai, DocSum.ai, and Atla AI} In this case study, we describe the document generation setting when an LLM is used by a lawyer or firm to draft a legal document. The input would be a prompt asking for a specific type of legal document (e.g., a contract or brief) for a specific purpose and the output would be the resulting document. 

In this setting, a lawyer may want \textbf{contributive} attributions to understand which training documents the generated document is borrowing words or concepts from. For example, if the document requested is a bespoke rental contract, users may want to ensure that the generated contract is not borrowing from rental contracts from other states or countries. Continuing with the rental contract example, \textbf{corroborative} attributions are also important to ensure the contract adheres to local laws. The sources for such corroborative attributions need not be in the training data and may come from a repository of documents that are more frequently updated than the language model itself.
In this setting, the LLM is assistive to lawyers handling the case. Correct attributions that provide the right sources to corroborate the drafted document are important. High-precision attributions in particular would improve the efficiency of lawyers using these tools. 

\subsection{Case Study 2: LLMs for Healthcare}

The application of language models to the field of medicine has been heavily studied \citep{singhal2022large, lewis-etal-2020-pretrained, Lee_2019, Luo_2022, Gu_2021, liévin2023large}. Recently, LLMs have been increasingly adopted for real-world clinical tasks that largely fall into the two categories of summarization of clinical notes \citep{abacha2023empirical,chuang2023spec,liu-etal-2023-deakinnlp} and medical QA \citep{cao2011askhermes, liévin2023large,singhal2023towards}. 


The task of summarizing clinical notes has received attention in both academia and industry.\footnote{Start-ups for summarization include Notable Health and Abridge AI.} These summaries have been evaluated for consistency with the underlying clinical notes using automated metrics, such as ROUGE and BERTScore \citep{zhang2020bertscore}, and human evaluators \citep{abacha2023empirical,chuang2023spec,liu-etal-2023-deakinnlp}.
While the \textbf{corroboration} of a generated summary with the sources it seeks to summarize
is critical, \textbf{contributive}
attributions could also be important in determining whether relevant training sources are influential. If training sources deemed irrelevant by domain knowledge are influential, then further precautions should be taken to monitor and improve the model.
Together, these attributions can provide insights into the validity of a summary of clinical notes.

For medical QA systems\footnote{Companies in this area include  MediSearch, Open Evidence, Hippocratic AI, and Glass Health.}, it is important for clinicians to have citations of evidence to support model answers \citep{sallam2023chatgpt}. \textbf{Corroborative} attributions can be used to provide these citations, as is done by MediSearch and OpenEvidence. While these two companies broadly restrict their attribution domains to research publications from reputable venues, MedAlign \citep{fleming2023medalign} highlights the option of using a corpus of EHRs. The implementation of corroborative attributions with trusted attribution domains is adjacent to the use case of fact checking, the stakes of which are particularly high in the clinical setting due to the potential consequences on human health.

Notions of attribution may also be valuable in debugging medical QA LLMs, such as MedPaLM 2~\citep{singhal2023towards}, by flagging training sources that are relevant to incorrect outputs. As discussed previously in \ref{corr_and_contr_use_cases}, this can be accomplished with a composition of \textbf{contributive} and \textbf{corroborative} attributions. Model developers and medical experts should leverage domain knowledge when manually inspecting training sources flagged for debugging.




\section{Future Work}
\label{sec:future work}
We highlight several promising directions for future work. 

\paragraph{Counterfactual contribution to output evaluators}
In Definition \ref{def:contribEval}, we outline the possibility of contributive evaluators that are sensitive to semantic changes in the counterfactual output, rather than to changes in the counterfactual loss. The notion of citation to parametric content discussed by Huang et al. \cite{huang2023citation} also addresses this potential connection between contributive attribution and the semantic content of the output. To the best of our knowledge, such output-based contributive attributions for LLMs have not yet been explored. Future work in addressing this challenging technical problem would allow for semantically meaningful contributive attributions. 

\paragraph{Contributive attributions with large-scale training data}
The large scale of data used to train LLMs raises concerns not only about the high resource burdens of TDA methods, but also whether the influence of a single training source is meaningfully noticeable on the loss, not to mention the output. Past work has quantitatively observed that training sources with high influences are more rare than not, but they do exist and in fact largely make up the total influence on an LLM output \citep{grosse2023studying}. Nonetheless, future work may consider extending contributive attributions for language models to notions of influence on a group of training sources, rather than individual training sources \citep{koh2019accuracy}. Also, the ubiquity of finetuning encourages further work on TDA methods suited for finetuned models \citep{kwon2023datainf}. In this case, the attribution domain could be restricted to the finetuning dataset, which is orders of magnitude smaller than the pre-training dataset. This direction is an interesting pursuit in and of itself, especially for model developers interested in debugging fine-tuned models.

\paragraph{Hybrid attribution systems}
While we present a framework that unifies existing work in both corroborative and contributive attribution literature, developing techniques capable of both types of attributions is left to future work. The area of \emph{fact-tracing} makes a step in this direction by providing contributive attributions in a setting where corroboration matters \citep{akyurek2022towards}. However, the identification and corroboration of facts within the language model output requires further work. Hybrid attribution systems would improve the customizability of attributions, potentially making them useful across a broader range of applications.  

\paragraph{Standardized Evaluation} From our survey of attribution methods, particularly for corroborative attribution, we observe that evaluation is not standardized between methods. Each attribution method is evaluated on different datasets and often with different metrics. For example, GopherCITE's~\citep{menick2022teaching} outputs are evaluated on a subset of  NaturalQuestions and ELI5 with binary metrics if the answer is plausible and supported by the attribution. On the other hand, WebGPT's~\citep{nakano2021webgpt} outputs are evaluated on a different subset of ELI5 and open-ended dialogue interactions by comparisons to human-generated attributions. More broadly, the utility of an attribution can be expanded beyond correctness to the other properties we introduce.

\paragraph{Use-Case Driven Method Development and Properties-Guided Evaluation} In our work, we explore tasks and case studies where attributions are important for industry applications of LLMs. We recommend that attribution system developers choose a use case and then identify the relevant properties for evaluation. This approach of goal-driven development is preferable to strong-arming a developed method to serve a use case. Furthermore, goal-driven development may surface additional settings where corroborative and contributive attributions are needed simultaneously.
\section{Conclusion}
\label{sec:conclusion}

This paper presents a unifying framework for corroborative and contributive attributions in LLMs. We formulate an interaction model to define the core components of attributions and to define their properties. This framework serves as a lens for analyzing existing attribution methods and use cases for attributions. Our analysis elucidates prescriptive suggestions for future research, namely CCO evaluators, the challenges of contributive methods at the scale of LLMs, the value of hybrid attributions systems, the need for standardized evaluation of attribution systems, and goal-driven development. We hope our unifying perspective on the field of attributions leads to improved solutions for misinformation, accountability, and transparency in real-world applications of language models.

\section{Acknowledgements}
This paper was developed in a fairness, accountability, transparency, and explainability working group run by Carlos Guestrin. We would like to thank Anka Reul, Ava Jeffs, Krista Opsahl-Ong, Myra Cheng, and Xuechen Li as well as all members of the working group for their thoughts and feedback in early discussions. We also thank Tatsunori Hashimoto and John Hewitt for their feedback on our manuscript. TW and NM are supported by the National Science Foundation Graduate Research Fellowship Program under Grant
No. DGE-2146755. Any opinion, findings, and conclusions or recommendations expressed in this material are those of the authors(s) and do not necessarily reflect the views of the National Science Foundation. NM was also supported by the Stanford Electrical Engineering Department Fellowship. JHS is supported by the Simons collaboration on the theory of algorithmic fairness and the Simons Foundation Investigators award 689988. CG is a Chan Zuckerberg Biohub – San Francisco Investigator. The figures in this work have been designed using images from Flaticon.


\bibliographystyle{IEEEtran}  
\bibliography{references} 

\begin{thebibliography}{10}
\providecommand{\url}[1]{#1}
\csname url@samestyle\endcsname
\providecommand{\newblock}{\relax}
\providecommand{\bibinfo}[2]{#2}
\providecommand{\BIBentrySTDinterwordspacing}{\spaceskip=0pt\relax}
\providecommand{\BIBentryALTinterwordstretchfactor}{4}
\providecommand{\BIBentryALTinterwordspacing}{\spaceskip=\fontdimen2\font plus
\BIBentryALTinterwordstretchfactor\fontdimen3\font minus
  \fontdimen4\font\relax}
\providecommand{\BIBforeignlanguage}[2]{{%
\expandafter\ifx\csname l@#1\endcsname\relax
\typeout{** WARNING: IEEEtran.bst: No hyphenation pattern has been}%
\typeout{** loaded for the language `#1'. Using the pattern for}%
\typeout{** the default language instead.}%
\else
\language=\csname l@#1\endcsname
\fi
#2}}
\providecommand{\BIBdecl}{\relax}
\BIBdecl

\bibitem{azamfirei2023large}
R.~Azamfirei, S.~R. Kudchadkar, and J.~Fackler, ``Large language models and the
  perils of their hallucinations,'' \emph{Critical Care}, vol.~27, no.~1, pp.
  1--2, 2023.

\bibitem{bommasani2021opportunities}
R.~Bommasani, D.~A. Hudson, E.~Adeli, R.~Altman, S.~Arora, S.~von Arx, M.~S.
  Bernstein, J.~Bohg, A.~Bosselut, E.~Brunskill \emph{et~al.}, ``On the
  opportunities and risks of foundation models,'' \emph{arXiv preprint
  arXiv:2108.07258}, 2021.

\bibitem{yue2023automatic}
X.~Yue, B.~Wang, K.~Zhang, Z.~Chen, Y.~Su, and H.~Sun, ``Automatic evaluation
  of attribution by large language models,'' \emph{arXiv preprint
  arXiv:2305.06311}, 2023.

\bibitem{guu2020retrieval}
K.~Guu, K.~Lee, Z.~Tung, P.~Pasupat, and M.~Chang, ``Retrieval augmented
  language model pre-training,'' in \emph{International conference on machine
  learning}.\hskip 1em plus 0.5em minus 0.4em\relax PMLR, 2020, pp. 3929--3938.

\bibitem{gao2023enabling}
T.~Gao, H.~Yen, J.~Yu, and D.~Chen, ``Enabling large language models to
  generate text with citations,'' 2023.

\bibitem{bohnet2022attributed}
B.~Bohnet, V.~Q. Tran, P.~Verga, R.~Aharoni, D.~Andor, L.~B. Soares,
  J.~Eisenstein, K.~Ganchev, J.~Herzig, K.~Hui \emph{et~al.}, ``Attributed
  question answering: Evaluation and modeling for attributed large language
  models,'' \emph{arXiv preprint arXiv:2212.08037}, 2022.

\bibitem{rashkin2021measuring}
H.~Rashkin, V.~Nikolaev, M.~Lamm, L.~Aroyo, M.~Collins, D.~Das, S.~Petrov,
  G.~S. Tomar, I.~Turc, and D.~Reitter, ``Measuring attribution in natural
  language generation models,'' \emph{arXiv preprint arXiv:2112.12870}, 2021.

\bibitem{koh2017understanding}
P.~W. Koh and P.~Liang, ``Understanding black-box predictions via influence
  functions,'' in \emph{International conference on machine learning}.\hskip
  1em plus 0.5em minus 0.4em\relax PMLR, 2017, pp. 1885--1894.

\bibitem{guu2023simfluence}
K.~Guu, A.~Webson, E.~Pavlick, L.~Dixon, I.~Tenney, and T.~Bolukbasi,
  ``Simfluence: Modeling the influence of individual training examples by
  simulating training runs,'' 2023.

\bibitem{ilyas2022datamodels}
A.~Ilyas, S.~M. Park, L.~Engstrom, G.~Leclerc, and A.~Madry, ``Datamodels:
  Predicting predictions from training data,'' \emph{arXiv preprint
  arXiv:2202.00622}, 2022.

\bibitem{yeh2018representer}
C.-K. Yeh, J.~S. Kim, I.~E.~H. Yen, and P.~Ravikumar, ``Representer point
  selection for explaining deep neural networks,'' 2018.

\bibitem{pruthi2020estimating}
G.~Pruthi, F.~Liu, S.~Kale, and M.~Sundararajan, ``Estimating training data
  influence by tracing gradient descent,'' \emph{Advances in Neural Information
  Processing Systems}, vol.~33, pp. 19\,920--19\,930, 2020.

\bibitem{schioppa2021scaling}
A.~Schioppa, P.~Zablotskaia, D.~Vilar, and A.~Sokolov, ``Scaling up influence
  functions,'' 2021.

\bibitem{kwon2023datainf}
Y.~Kwon, E.~Wu, K.~Wu, and J.~Zou, ``Datainf: Efficiently estimating data
  influence in lora-tuned llms and diffusion models,'' 2023.

\bibitem{grosse2023studying}
R.~Grosse, J.~Bae, C.~Anil, N.~Elhage, A.~Tamkin, A.~Tajdini, B.~Steiner,
  D.~Li, E.~Durmus, E.~Perez \emph{et~al.}, ``Studying large language model
  generalization with influence functions,'' \emph{arXiv preprint
  arXiv:2308.03296}, 2023.

\bibitem{park2023trak}
S.~M. Park, K.~Georgiev, A.~Ilyas, G.~Leclerc, and A.~Madry, ``Trak:
  Attributing model behavior at scale,'' 2023.

\bibitem{huang2023citation}
J.~Huang and K.~C.-C. Chang, ``Citation: A key to building responsible and
  accountable large language models,'' 2023.

\bibitem{liu2023evaluating}
N.~F. Liu, T.~Zhang, and P.~Liang, ``Evaluating verifiability in generative
  search engines,'' 2023.

\bibitem{lundberg2017unified}
S.~Lundberg and S.-I. Lee, ``A unified approach to interpreting model
  predictions,'' 2017.

\bibitem{akyurek2022towards}
E.~Aky{\"u}rek, T.~Bolukbasi, F.~Liu, B.~Xiong, I.~Tenney, J.~Andreas, and
  K.~Guu, ``Towards tracing knowledge in language models back to the training
  data,'' in \emph{Findings of the Association for Computational Linguistics:
  EMNLP 2022}, 2022, pp. 2429--2446.

\bibitem{honovich2022true}
O.~Honovich, R.~Aharoni, J.~Herzig, H.~Taitelbaum, D.~Kukliansy, V.~Cohen,
  T.~Scialom, I.~Szpektor, A.~Hassidim, and Y.~Matias, ``True: Re-evaluating
  factual consistency evaluation,'' \emph{arXiv preprint arXiv:2204.04991},
  2022.

\bibitem{gao2023rarr}
L.~Gao, Z.~Dai, P.~Pasupat, A.~Chen, A.~T. Chaganty, Y.~Fan, V.~Y. Zhao,
  N.~Lao, H.~Lee, D.-C. Juan, and K.~Guu, ``Rarr: Researching and revising what
  language models say, using language models,'' 2023.

\bibitem{hammoudeh2022training}
Z.~Hammoudeh and D.~Lowd, ``Training data influence analysis and estimation: A
  survey,'' \emph{arXiv preprint arXiv:2212.04612}, 2022.

\bibitem{madsen2022post}
A.~Madsen, S.~Reddy, and S.~Chandar, ``Post-hoc interpretability for neural
  nlp: A survey,'' \emph{ACM Computing Surveys}, vol.~55, no.~8, pp. 1--42,
  2022.

\bibitem{ramos2003using}
J.~Ramos \emph{et~al.}, ``Using tf-idf to determine word relevance in document
  queries,'' in \emph{Proceedings of the first instructional conference on
  machine learning}, vol. 242, no.~1.\hskip 1em plus 0.5em minus 0.4em\relax
  Citeseer, 2003, pp. 29--48.

\bibitem{roberts-etal-2020-much}
A.~Roberts, C.~Raffel, and N.~Shazeer, ``How much knowledge can you pack into
  the parameters of a language model?'' in \emph{Proceedings of the 2020
  Conference on Empirical Methods in Natural Language Processing
  (EMNLP)}.\hskip 1em plus 0.5em minus 0.4em\relax Online: Association for
  Computational Linguistics, Nov. 2020.

\bibitem{nakamura-etal-2022-hybridialogue}
K.~Nakamura, S.~Levy, Y.-L. Tuan, W.~Chen, and W.~Y. Wang, ``Hybridialogue: An
  information-seeking dialogue dataset grounded on tabular and textual data,''
  2022.

\bibitem{menick2022teaching}
J.~Menick, M.~Trebacz, V.~Mikulik, J.~Aslanides, F.~Song, M.~Chadwick,
  M.~Glaese, S.~Young, L.~Campbell-Gillingham, G.~Irving, and N.~McAleese,
  ``Teaching language models to support answers with verified quotes,'' 2022.

\bibitem{cosijn2000dimensions}
E.~Cosijn and P.~Ingwersen, ``Dimensions of relevance,'' \emph{Information
  Processing \& Management}, vol.~36, no.~4, pp. 533--550, 2000.

\bibitem{guo2022survey}
Z.~Guo, M.~Schlichtkrull, and A.~Vlachos, ``A survey on automated
  fact-checking,'' \emph{Transactions of the Association for Computational
  Linguistics}, vol.~10, pp. 178--206, 2022.

\bibitem{yue2022c}
X.~Yue, X.~Pan, W.~Yao, D.~Yu, D.~Yu, and J.~Chen, ``C-more: Pretraining to
  answer open-domain questions by consulting millions of references,''
  \emph{arXiv preprint arXiv:2203.08928}, 2022.

\bibitem{zadeh1975concept}
L.~A. Zadeh, ``The concept of a linguistic variable and its application to
  approximate reasoning—i,'' \emph{Information sciences}, vol.~8, no.~3, pp.
  199--249, 1975.

\bibitem{erasmus2021interpretability}
A.~Erasmus, T.~D. Brunet, and E.~Fisher, ``What is interpretability?''
  \emph{Philosophy \& Technology}, vol.~34, no.~4, pp. 833--862, 2021.

\bibitem{yeh2022first}
C.-K. Yeh, A.~Taly, M.~Sundararajan, F.~Liu, and P.~Ravikumar, ``First is
  better than last for language data influence,'' \emph{Advances in Neural
  Information Processing Systems}, vol.~35, pp. 32\,285--32\,298, 2022.

\bibitem{basu2020influence}
S.~Basu, P.~Pope, and S.~Feizi, ``Influence functions in deep learning are
  fragile,'' \emph{arXiv preprint arXiv:2006.14651}, 2020.

\bibitem{sogaard2021revisiting}
A.~S{\o}gaard \emph{et~al.}, ``Revisiting methods for finding influential
  examples,'' \emph{arXiv preprint arXiv:2111.04683}, 2021.

\bibitem{bender2021dangers}
E.~M. Bender, T.~Gebru, A.~McMillan-Major, and S.~Shmitchell, ``On the dangers
  of stochastic parrots: Can language models be too big?'' in \emph{Proceedings
  of the 2021 ACM conference on fairness, accountability, and transparency},
  2021, pp. 610--623.

\bibitem{liang2022holistic}
P.~Liang, R.~Bommasani, T.~Lee, D.~Tsipras, D.~Soylu, M.~Yasunaga, Y.~Zhang,
  D.~Narayanan, Y.~Wu, A.~Kumar \emph{et~al.}, ``Holistic evaluation of
  language models,'' \emph{arXiv preprint arXiv:2211.09110}, 2022.

\bibitem{piktus2022web}
A.~Piktus, F.~Petroni, V.~Karpukhin, D.~Okhonko, S.~Broscheit, G.~Izacard,
  P.~Lewis, B.~Oğuz, E.~Grave, W.~tau Yih, and S.~Riedel, ``The web is your
  oyster - knowledge-intensive nlp against a very large web corpus,'' 2022.

\bibitem{ni-etal-2022-large}
J.~Ni, C.~Qu, J.~Lu, Z.~Dai, G.~Hernandez~Abrego, J.~Ma, V.~Zhao, Y.~Luan,
  K.~Hall, M.-W. Chang, and Y.~Yang, ``Large dual encoders are generalizable
  retrievers,'' in \emph{Proceedings of the 2022 Conference on Empirical
  Methods in Natural Language Processing}.\hskip 1em plus 0.5em minus
  0.4em\relax Abu Dhabi, United Arab Emirates: Association for Computational
  Linguistics, Dec. 2022.

\bibitem{karpukhin-etal-2020-dense}
V.~Karpukhin, B.~Oguz, S.~Min, P.~Lewis, L.~Wu, S.~Edunov, D.~Chen, and W.-t.
  Yih, ``Dense passage retrieval for open-domain question answering,'' in
  \emph{Proceedings of the 2020 Conference on Empirical Methods in Natural
  Language Processing (EMNLP)}.\hskip 1em plus 0.5em minus 0.4em\relax Online:
  Association for Computational Linguistics, Nov. 2020.

\bibitem{robertson2009probabilistic}
S.~Robertson, H.~Zaragoza \emph{et~al.}, ``The probabilistic relevance
  framework: Bm25 and beyond,'' \emph{Foundations and Trends in Information
  Retrieval}, vol.~3, no.~4, pp. 333--389, 2009.

\bibitem{rae2022scaling}
J.~W. Rae, S.~Borgeaud, T.~Cai, K.~Millican, J.~Hoffmann, F.~Song,
  J.~Aslanides, S.~Henderson, R.~Ring, S.~Young, E.~Rutherford, T.~Hennigan,
  J.~Menick, A.~Cassirer, R.~Powell, G.~van~den Driessche, L.~A. Hendricks,
  M.~Rauh, P.-S. Huang, A.~Glaese, J.~Welbl, S.~Dathathri, S.~Huang, J.~Uesato,
  J.~Mellor, I.~Higgins, A.~Creswell, N.~McAleese, A.~Wu, E.~Elsen,
  S.~Jayakumar, E.~Buchatskaya, D.~Budden, E.~Sutherland, K.~Simonyan,
  M.~Paganini, L.~Sifre, L.~Martens, X.~L. Li, A.~Kuncoro, A.~Nematzadeh,
  E.~Gribovskaya, D.~Donato, A.~Lazaridou, A.~Mensch, J.-B. Lespiau,
  M.~Tsimpoukelli, N.~Grigorev, D.~Fritz, T.~Sottiaux, M.~Pajarskas, T.~Pohlen,
  Z.~Gong, D.~Toyama, C.~de~Masson~d'Autume, Y.~Li, T.~Terzi, V.~Mikulik,
  I.~Babuschkin, A.~Clark, D.~de~Las~Casas, A.~Guy, C.~Jones, J.~Bradbury,
  M.~Johnson, B.~Hechtman, L.~Weidinger, I.~Gabriel, W.~Isaac, E.~Lockhart,
  S.~Osindero, L.~Rimell, C.~Dyer, O.~Vinyals, K.~Ayoub, J.~Stanway,
  L.~Bennett, D.~Hassabis, K.~Kavukcuoglu, and G.~Irving, ``Scaling language
  models: Methods, analysis \& insights from training gopher,'' 2022.

\bibitem{thoppilan2022lamda}
R.~Thoppilan, D.~D. Freitas, J.~Hall, N.~Shazeer, A.~Kulshreshtha, H.-T. Cheng,
  A.~Jin, T.~Bos, L.~Baker, Y.~Du, Y.~Li, H.~Lee, H.~S. Zheng, A.~Ghafouri,
  M.~Menegali, Y.~Huang, M.~Krikun, D.~Lepikhin, J.~Qin, D.~Chen, Y.~Xu,
  Z.~Chen, A.~Roberts, M.~Bosma, V.~Zhao, Y.~Zhou, C.-C. Chang, I.~Krivokon,
  W.~Rusch, M.~Pickett, P.~Srinivasan, L.~Man, K.~Meier-Hellstern, M.~R.
  Morris, T.~Doshi, R.~D. Santos, T.~Duke, J.~Soraker, B.~Zevenbergen,
  V.~Prabhakaran, M.~Diaz, B.~Hutchinson, K.~Olson, A.~Molina, E.~Hoffman-John,
  J.~Lee, L.~Aroyo, R.~Rajakumar, A.~Butryna, M.~Lamm, V.~Kuzmina, J.~Fenton,
  A.~Cohen, R.~Bernstein, R.~Kurzweil, B.~Aguera-Arcas, C.~Cui, M.~Croak,
  E.~Chi, and Q.~Le, ``Lamda: Language models for dialog applications,'' 2022.

\bibitem{nakano2021webgpt}
R.~Nakano, J.~Hilton, S.~Balaji, J.~Wu, L.~Ouyang, C.~Kim, C.~Hesse, S.~Jain,
  V.~Kosaraju, W.~Saunders \emph{et~al.}, ``Webgpt: Browser-assisted
  question-answering with human feedback,'' \emph{arXiv preprint
  arXiv:2112.09332}, 2021.

\bibitem{lazaridou2022internet}
A.~Lazaridou, E.~Gribovskaya, W.~Stokowiec, and N.~Grigorev,
  ``Internet-augmented language models through few-shot prompting for
  open-domain question answering,'' \emph{arXiv preprint arXiv:2203.05115},
  2022.

\bibitem{chowdhery2022palm}
A.~Chowdhery, S.~Narang, J.~Devlin, M.~Bosma, G.~Mishra, A.~Roberts, P.~Barham,
  H.~W. Chung, C.~Sutton, S.~Gehrmann \emph{et~al.}, ``Palm: Scaling language
  modeling with pathways,'' \emph{arXiv preprint arXiv:2204.02311}, 2022.

\bibitem{wei2023chainofthought}
J.~Wei, X.~Wang, D.~Schuurmans, M.~Bosma, B.~Ichter, F.~Xia, E.~Chi, Q.~Le, and
  D.~Zhou, ``Chain-of-thought prompting elicits reasoning in large language
  models,'' 2023.

\bibitem{ghorbani2019data}
A.~Ghorbani and J.~Zou, ``Data shapley: Equitable valuation of data for machine
  learning,'' in \emph{International Conference on Machine Learning}.\hskip 1em
  plus 0.5em minus 0.4em\relax PMLR, 2019, pp. 2242--2251.

\bibitem{bae2022if}
J.~Bae, N.~Ng, A.~Lo, M.~Ghassemi, and R.~B. Grosse, ``If influence functions
  are the answer, then what is the question?'' \emph{Advances in Neural
  Information Processing Systems}, vol.~35, pp. 17\,953--17\,967, 2022.

\bibitem{pmlr-v97-ghorbani19c}
A.~Ghorbani and J.~Zou, ``Data shapley: Equitable valuation of data for machine
  learning,'' 2019.

\bibitem{choudhury2023investigating}
A.~Choudhury and H.~Shamszare, ``Investigating the impact of user trust on the
  adoption and use of chatgpt: Survey analysis,'' \emph{Journal of Medical
  Internet Research}, vol.~25, p. e47184, 2023.

\bibitem{khan2023unstoppable}
U.~A. Khan, ``The unstoppable march of artificial intelligence: The dawn of
  large language models,'' \emph{eSignals PRO}, 2023.

\bibitem{krause2020fact}
N.~M. Krause, I.~Freiling, B.~Beets, and D.~Brossard, ``Fact-checking as risk
  communication: the multi-layered risk of misinformation in times of
  covid-19,'' \emph{Journal of Risk Research}, vol.~23, no. 7-8, pp.
  1052--1059, 2020.

\bibitem{amazeen2015revisiting}
M.~A. Amazeen, ``Revisiting the epistemology of fact-checking,'' \emph{Critical
  Review}, vol.~27, no.~1, pp. 1--22, 2015.

\bibitem{samuelson2023generative}
P.~Samuelson, ``Generative ai meets copyright,'' \emph{Science}, vol. 381, no.
  6654, pp. 158--161, 2023.

\bibitem{brophy2020exit}
J.~Brophy, ``Exit through the training data: A look into instance-attribution
  explanations and efficient data deletion in machine learning,''
  \emph{Technical report}, 2020.

\bibitem{feldman2020neural}
V.~Feldman and C.~Zhang, ``What neural networks memorize and why: Discovering
  the long tail via influence estimation,'' 2020.

\bibitem{sun2022investigating}
J.~Sun, Q.~V. Liao, M.~Muller, M.~Agarwal, S.~Houde, K.~Talamadupula, and J.~D.
  Weisz, ``Investigating explainability of generative ai for code through
  scenario-based design,'' in \emph{27th International Conference on
  Intelligent User Interfaces}, 2022, pp. 212--228.

\bibitem{liao2020questioning}
Q.~V. Liao, D.~Gruen, and S.~Miller, ``Questioning the ai: informing design
  practices for explainable ai user experiences,'' in \emph{Proceedings of the
  2020 CHI conference on human factors in computing systems}, 2020, pp. 1--15.

\bibitem{hendersonkrass2022pileoflaw}
P.~Henderson, M.~S. Krass, L.~Zheng, N.~Guha, C.~D. Manning, D.~Jurafsky, and
  D.~E. Ho, ``Pile of law: Learning responsible data filtering from the law and
  a 256gb open-source legal dataset,'' 2022.

\bibitem{guha2023legalbench}
N.~Guha, J.~Nyarko, D.~E. Ho, C.~Ré, A.~Chilton, A.~Narayana, A.~Chohlas-Wood,
  A.~Peters, B.~Waldon, D.~N. Rockmore, D.~Zambrano, D.~Talisman, E.~Hoque,
  F.~Surani, F.~Fagan, G.~Sarfaty, G.~M. Dickinson, H.~Porat, J.~Hegland,
  J.~Wu, J.~Nudell, J.~Niklaus, J.~Nay, J.~H. Choi, K.~Tobia, M.~Hagan, M.~Ma,
  M.~Livermore, N.~Rasumov-Rahe, N.~Holzenberger, N.~Kolt, P.~Henderson,
  S.~Rehaag, S.~Goel, S.~Gao, S.~Williams, S.~Gandhi, T.~Zur, V.~Iyer, and
  Z.~Li, ``Legalbench: A collaboratively built benchmark for measuring legal
  reasoning in large language models,'' 2023.

\bibitem{niklaus2023multilegalpile}
J.~Niklaus, V.~Matoshi, M.~Stürmer, I.~Chalkidis, and D.~E. Ho,
  ``Multilegalpile: A 689gb multilingual legal corpus,'' 2023.

\bibitem{cui2023chatlaw}
J.~Cui, Z.~Li, Y.~Yan, B.~Chen, and L.~Yuan, ``Chatlaw: Open-source legal large
  language model with integrated external knowledge bases,'' 2023.

\bibitem{shaghaghian2021customizing}
S.~Shaghaghian, Luna, Feng, B.~Jafarpour, and N.~Pogrebnyakov, ``Customizing
  contextualized language models forlegal document reviews,'' 2021.

\bibitem{nay2023large}
J.~J. Nay, D.~Karamardian, S.~B. Lawsky, W.~Tao, M.~Bhat, R.~Jain, A.~T. Lee,
  J.~H. Choi, and J.~Kasai, ``Large language models as tax attorneys: A case
  study in legal capabilities emergence,'' 2023.

\bibitem{sun2023short}
Z.~Sun, ``A short survey of viewing large language models in legal aspect,''
  2023.

\bibitem{deroy2023ready}
A.~Deroy, K.~Ghosh, and S.~Ghosh, ``How ready are pre-trained abstractive
  models and llms for legal case judgement summarization?'' 2023.

\bibitem{singhal2022large}
K.~Singhal, S.~Azizi, T.~Tu, S.~S. Mahdavi, J.~Wei, H.~W. Chung, N.~Scales,
  A.~Tanwani, H.~Cole-Lewis, S.~Pfohl, P.~Payne, M.~Seneviratne, P.~Gamble,
  C.~Kelly, N.~Scharli, A.~Chowdhery, P.~Mansfield, B.~A. y~Arcas, D.~Webster,
  G.~S. Corrado, Y.~Matias, K.~Chou, J.~Gottweis, N.~Tomasev, Y.~Liu,
  A.~Rajkomar, J.~Barral, C.~Semturs, A.~Karthikesalingam, and V.~Natarajan,
  ``Large language models encode clinical knowledge,'' 2022.

\bibitem{lewis-etal-2020-pretrained}
P.~Lewis, M.~Ott, J.~Du, and V.~Stoyanov, ``Pretrained language models for
  biomedical and clinical tasks: Understanding and extending the
  state-of-the-art,'' in \emph{Proceedings of the 3rd Clinical Natural Language
  Processing Workshop}.\hskip 1em plus 0.5em minus 0.4em\relax Online:
  Association for Computational Linguistics, Nov. 2020.

\bibitem{Lee_2019}
J.~Lee, W.~Yoon, S.~Kim, D.~Kim, S.~Kim, C.~H. So, and J.~Kang, ``Biobert: a
  pre-trained biomedical language representation model for biomedical text
  mining,'' \emph{Bioinformatics}, vol.~36, no.~4, p. 1234–1240, Sep. 2019.

\bibitem{Luo_2022}
R.~Luo, L.~Sun, Y.~Xia, T.~Qin, S.~Zhang, H.~Poon, and T.-Y. Liu, ``Biogpt:
  generative pre-trained transformer for biomedical text generation and
  mining,'' \emph{Briefings in Bioinformatics}, vol.~23, no.~6, Sep. 2022.

\bibitem{Gu_2021}
Y.~Gu, R.~Tinn, H.~Cheng, M.~Lucas, N.~Usuyama, X.~Liu, T.~Naumann, J.~Gao, and
  H.~Poon, ``Domain-specific language model pretraining for biomedical natural
  language processing,'' \emph{{ACM} Transactions on Computing for Healthcare},
  vol.~3, no.~1, pp. 1--23, oct 2021.

\bibitem{liévin2023large}
V.~Liévin, C.~E. Hother, and O.~Winther, ``Can large language models reason
  about medical questions?'' 2023.

\bibitem{abacha2023empirical}
A.~B. Abacha, W.-w. Yim, Y.~Fan, and T.~Lin, ``An empirical study of clinical
  note generation from doctor-patient encounters,'' in \emph{Proceedings of the
  17th Conference of the European Chapter of the Association for Computational
  Linguistics}, 2023, pp. 2283--2294.

\bibitem{chuang2023spec}
Y.-N. Chuang, R.~Tang, X.~Jiang, and X.~Hu, ``Spec: A soft prompt-based
  calibration on performance variability of large language model in clinical
  notes summarization,'' 2023.

\bibitem{liu-etal-2023-deakinnlp}
M.~Liu, D.~Zhang, W.~Tan, and H.~Zhang, ``{D}eakin{NLP} at {P}rob{S}um 2023:
  Clinical progress note summarization with rules and language
  {M}odels{C}linical progress note summarization with rules and languague
  models,'' in \emph{The 22nd Workshop on Biomedical Natural Language
  Processing and BioNLP Shared Tasks}.\hskip 1em plus 0.5em minus 0.4em\relax
  Toronto, Canada: Association for Computational Linguistics, Jul. 2023.

\bibitem{cao2011askhermes}
Y.~Cao, F.~Liu, P.~Simpson, L.~Antieau, A.~Bennett, J.~J. Cimino, J.~Ely, and
  H.~Yu, ``Askhermes: An online question answering system for complex clinical
  questions,'' \emph{Journal of biomedical informatics}, vol.~44, no.~2, pp.
  277--288, 2011.

\bibitem{singhal2023towards}
K.~Singhal, T.~Tu, J.~Gottweis, R.~Sayres, E.~Wulczyn, L.~Hou, K.~Clark,
  S.~Pfohl, H.~Cole-Lewis, D.~Neal \emph{et~al.}, ``Towards expert-level
  medical question answering with large language models,'' \emph{arXiv preprint
  arXiv:2305.09617}, 2023.

\bibitem{zhang2020bertscore}
T.~Zhang, V.~Kishore, F.~Wu, K.~Q. Weinberger, and Y.~Artzi, ``Bertscore:
  Evaluating text generation with bert,'' 2020.

\bibitem{sallam2023chatgpt}
M.~Sallam, ``Chatgpt utility in healthcare education, research, and practice:
  systematic review on the promising perspectives and valid concerns,'' in
  \emph{Healthcare}, vol.~11, no.~6.\hskip 1em plus 0.5em minus 0.4em\relax
  MDPI, 2023, p. 887.

\bibitem{fleming2023medalign}
S.~L. Fleming, A.~Lozano, W.~J. Haberkorn, J.~A. Jindal, E.~P. Reis, R.~Thapa,
  L.~Blankemeier, J.~Z. Genkins, E.~Steinberg, A.~Nayak \emph{et~al.},
  ``Medalign: A clinician-generated dataset for instruction following with
  electronic medical records,'' \emph{arXiv preprint arXiv:2308.14089}, 2023.

\bibitem{koh2019accuracy}
P.~W.~W. Koh, K.-S. Ang, H.~Teo, and P.~S. Liang, ``On the accuracy of
  influence functions for measuring group effects,'' \emph{Advances in neural
  information processing systems}, vol.~32, 2019.

\bibitem{zhang2022locally}
S.~Zhang, J.~Wang, H.~Jiang, and R.~Song, ``Locally aggregated feature
  attribution on natural language model understanding,'' \emph{arXiv preprint
  arXiv:2204.10893}, 2022.

\end{thebibliography}
\newpage
\begin{appendices}
\section{In-context data as the attribution domain}
\label{in context}
Due to the ubiquity of prompt engineering techniques, data provided in-context is a highly relevant attribution domain.\\

\textbf{Defining sources within the attribution domain:} Some forms of in-context data, such as documents retrieved from an external corpora and few-shot examples, contain natural structure to determine the segments that correspond to individual sources. Other forms of in-context data, however, may not have such structure for delineating the boundaries between sources. In order to designate such forms of in-context data as an attribution domain under our framework, it is necessary for a task designer to mark the segments of the in-context data that correspond to individual sources, according to the specifics of the task at hand. 

For example, consider seeking contributive attributions to parts of the in-context data to gain insight into model behavior. Prior work refers to this setting as feature attribution \citep{zhang2022locally}, where each word of the in-context data is treated as an individual feature, or source. Here, we examine alternatives to defining each word of the input as a source. Consider the following examples of inputs, each containing different forms of in-context data, and their corresponding sources:  
\begin{enumerate}
    \item \textbf{Input with natural structure:} "Sentence: The moonlight gently illuminated the peaceful meadow. \\
    Sentiment: Positive \\
    Sentence: The sun cast harsh rays over the sweltering sand. \\
    Sentiment: Negative \\
    Sentence: The moonlight shone bright over the sparkling water. \\
    Sentiment:"
        \begin{enumerate}[start=0, label={\bfseries $s_\arabic*$:}]
            \item "Sentence: The moonlight gently illuminated the peaceful meadow. \\
    Sentiment: Positive"
            \item "Sentence: The sun cast harsh rays over the sweltering sand. \\
    Sentiment: Negative"
        \end{enumerate}
    \item \textbf{Input without natural structure:} "What are the lunar phases? What is a lunar eclipse? Explain like I'm five."
        \begin{enumerate}[start=0, label={\bfseries $s_\arabic*$:}]
            \item "What are the lunar phases? What is a lunar eclipse?"
            \item "Explain like I'm five."
        \end{enumerate}
\end{enumerate}

\textbf{Contributive attributions to in-context data:} The contributive evaluators defined in Definition \ref{def:contribEval} pertain to sources on which the model was trained, either during pre-training or fine-tuning; they all pose the counterfactual scenario of a model trained without the source in question. If, however, the attribution domain is data provided in-context, rather than during pre-training or fine-tuning, one could pose an alternative counterfactual: what would the model output be if the source had not been included in-context? With this alternative counterfactual in hand, the contributive evaluators discussed in Definition \ref{def:contribEval} could be extended to provide contributive attributions to an in-context attribution domain.\\

Our discussion so far presents one view of in-context data attribution through the lens of feature attribution. We hope future work will develop various paradigms and accompanying methods for generating and verifying in-context data attributions. 

\section{Clause-level Explicatures}
\label{cle}
The following definitions formally define clause-level explicatures, which can be used as attributable units for corroborative attributions.


\begin{definition} \textbf{Clause-level Standalone Proposition}. A standalone proposition, as defined by \citep{rashkin2021measuring}, that cannot be broken down into two or more non-overlapping standalone propositions.
\end{definition}

Consider the following examples:\\
\textbf{Example 1:} In 2010, Barack Obama was the president of the United States and a father of two.\\
\textbf{Example 2:} In 2010, Barack Obama was the president of the United States.\\
\textbf{Example 3:} In 2010, Barack Obama was a father of two.\\
\textbf{Example 4:} In 2010, Barack Obama was a father.\\
\\
Example 1 is a standalone proposition, but is not clause-level. Examples 2 and 3 are the clause-level standalone propositions that compose Example 1. Example 4 is also a clause-level standalone proposition. Note that example 3 contains the example 4 but they are overlapping because they share the same information; this does not prevent example 3 from being clause-level. 


\begin{definition} \textbf{Clause-level Explicature}. The clause-level standalone propositions contained within a sentence-level explicature, as defined by \citep{rashkin2021measuring}.
\end{definition}
A clause-level explicature is a clause-level standalone proposition that is fully interpretable given only the wall clock time at which the input was used to query the model. We refer readers to \citep{rashkin2021measuring} for the formal definition of a sentence-level explicature, which Definition \ref{cle} extends.



\section{Entrepreneurial motivation for LLM Attributions: Y-Combinator Case Study}
\label{yc}

Our motivation for introducing this unified framework of attributions is driven by the rapidly advancing development of large language models to increasingly high-stakes domains. To understand how LLMs will likely be used in the near future, we examine ventures that have been proposed and funded based on LLM technology. As a case study, we look through the Summer 2023 Y-Combinator class\footnote{https://www.ycombinator.com/blog/meet-the-yc-summer-2023-batch} and examine the ventures that use LLMs, and highlight where attributions, both corroborative and contributive, may be important. Of the 46 companies listed under and \textit{Generative AI}, 41 companies described the usage of large language models in various application domains (Table \ref{tab:yc}). The use cases (Section \ref{sec:applications}) and case studies (Section \ref{sec:case studies}) we study in our work are motivated by the different ways these companies have chosen to apply LLMs. Moreover, both corroborative attributions and contributive attributions may be helpful as these ventures and many others begin deployment LLMs in the real world. 

\begin{table}[]
\center
\begin{tabular}{l|l}
          Application Area      & Count \\ \hline
          Code Generation       & 12    \\
          AI Tools              & 9     \\
          Content Creation      & 8     \\
          Persona/Assistants/QA & 6     \\
          Healthcare            & 2     \\
          Legal                 & 3     \\
          Education             & 1     \\
          Total             &  41   
\end{tabular}
\caption{Summary of LLM application domains targeted by Generative AI ventures of the summer 2023 Y-Combinator class. 41/46 generative AI category companies included LLM outputs as part of their product or service.}
\label{tab:yc}
\end{table}
\end{appendices}

\end{document}